\documentclass{article}

\usepackage{times}
\usepackage{graphicx} 
\usepackage{subfigure}

\usepackage{natbib}

\usepackage{algorithm}
\usepackage{algorithmic}

\usepackage{url}
\usepackage{amsmath}
\usepackage{graphicx}
\usepackage{verbatim}
\usepackage{stmaryrd}

\usepackage{hyperref}

\long\def\ignorethis#1{}

\newcommand{\tr}{^\mathrm{T}}

\newcommand{\gauss}{\mathcal{N}}

\renewcommand{\eqref}[1]{Equation~(\ref{#1})}

\newcommand{\params}{\theta}

\newcommand{\cost}{c}
\newcommand{\monoreg}{g_{\operatorname{mono}}} 
\newcommand{\smoothreg}{g_{\operatorname{lcr}}} 
\newcommand{\state}{\mathbf{x}}
\newcommand{\action}{\mathbf{u}}

\newcommand{\traj}{\tau}

\newcommand{\obj}{\mathcal{L}}

\newcommand{\kl}{D_\text{KL}}
\newcommand{\ent}{\mathcal{H}}

\newcommand{\st}{\state_t}
\newcommand{\at}{\action_t}

\usepackage[accepted]{icml2016}

\icmltitlerunning{Guided Cost Learning}

\begin{document}

\twocolumn[
\icmltitle{Guided Cost Learning: Deep Inverse Optimal Control via Policy Optimization}

\icmlauthor{Chelsea Finn}{cbfinn@eecs.berkeley.edu}
\icmlauthor{Sergey Levine}{svlevine@eecs.berkeley.edu}
\icmlauthor{Pieter Abbeel}{pabbeel@eecs.berkeley.edu}
\icmladdress{University of California, Berkeley,
            Berkeley, CA 94709 USA}

\icmlkeywords{inverse optimal control, robotics}

\vskip 0.3in
]

\begin{abstract}
Reinforcement learning can acquire complex behaviors from high-level specifications. However, defining a cost function that can be optimized effectively and encodes the correct task is challenging in practice. We explore how inverse optimal control (IOC) can be used to learn behaviors from demonstrations, with applications to torque control of high-dimensional robotic systems. Our method addresses two key challenges in inverse optimal control: first, the need for informative features and effective regularization to impose structure on the cost, and second, the difficulty of learning the cost function under unknown dynamics for high-dimensional continuous systems. To address the former challenge, we present an algorithm capable of learning arbitrary nonlinear cost functions, such as neural networks, without meticulous feature engineering. To address the latter challenge, we formulate an efficient sample-based approximation for MaxEnt IOC.
We evaluate our method on a series of simulated tasks and real-world robotic manipulation problems, demonstrating substantial improvement over prior methods both in terms of task complexity and sample efficiency.
\end{abstract}

\section{Introduction}
\vspace{-.05cm}


Reinforcement learning can be used to acquire complex behaviors from high-level specifications.
However, defining a cost function that can be optimized effectively and encodes the correct task can be challenging in practice, and techniques like cost shaping are often used to solve complex real-world problems \cite{nhr-tars-99}.
Inverse optimal control (IOC) or inverse reinforcement learning (IRL) provide an avenue for addressing this challenge by learning a cost function directly from expert demonstrations, e.g. \citet{nr-airl-00,an-alirl-04,zmbd-meirl-08}.
However, designing an effective IOC algorithm for learning from demonstration is difficult for two reasons. First, IOC is fundamentally underdefined in that many costs induce the same behavior.
Most practical algorithms therefore require carefully designed features to impose structure on the learned cost. Second, many standard IRL and IOC methods require solving the forward problem (finding an optimal policy given the current cost) in the inner loop of an iterative cost optimization.
This makes them difficult to apply to complex, high-dimensional systems, where the forward problem is itself exceedingly difficult, particularly real-world robotic systems with unknown dynamics.

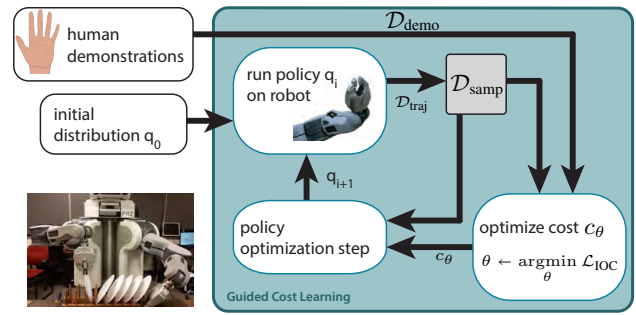
\begin{figure}
\setlength{\unitlength}{0.5\columnwidth}
\begin{picture}(1.99,0.92) \linethickness{0.5pt}

\put(-0.01,0.01){\includegraphics[width=1.01\columnwidth]{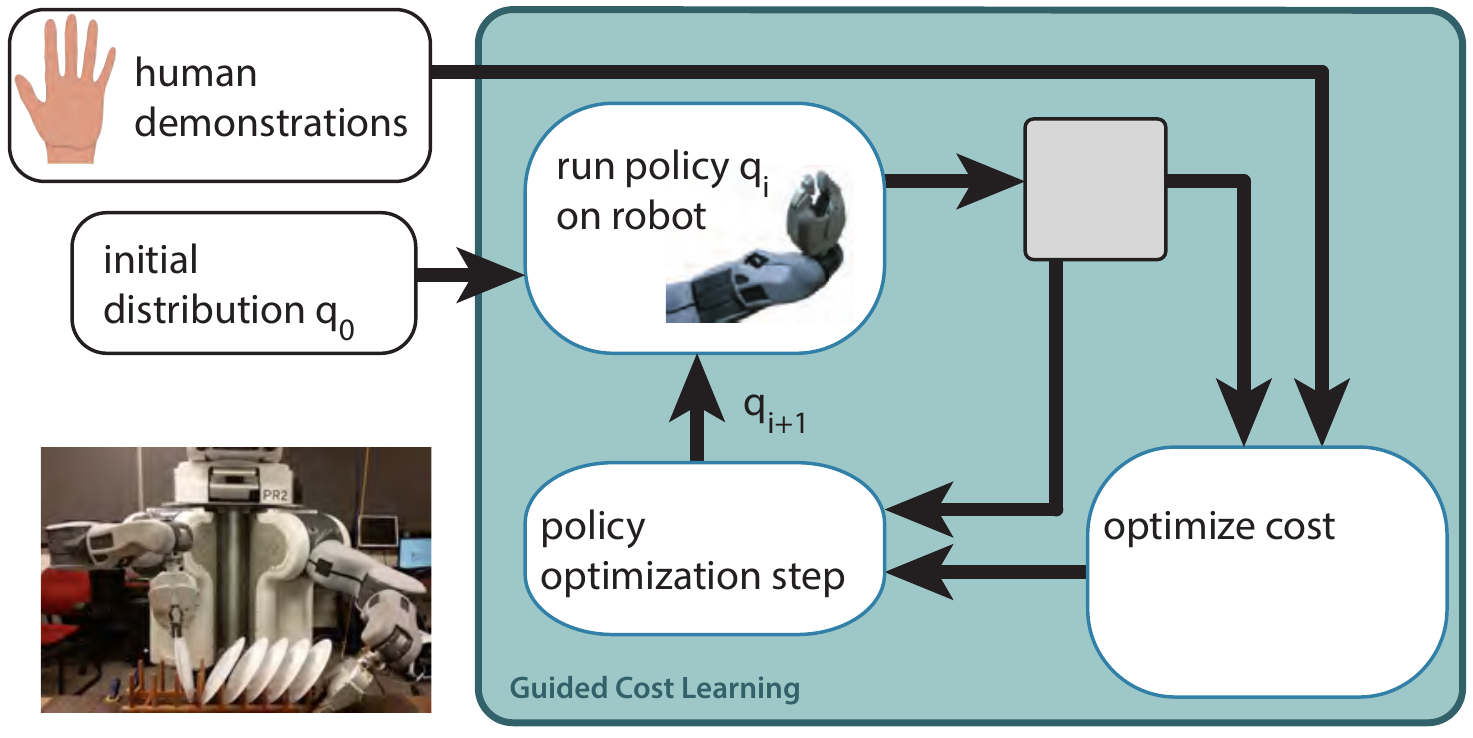}}
\put(1.2,0.945){\small{$\mathcal{D}_\text{demo}$}}

\put(1.41,0.74){\small{$\mathcal{D}_\text{samp}$}}

\put(1.23,0.69){\tiny{$\mathcal{D}_{\text{traj}}$}}

\put(1.36,0.195){\tiny{$\cost_\params$}}

\put(1.84,0.279){\small{$\cost_\params$}}

\put(1.51,0.175){\tiny{$\params \shortleftarrow \underset{\params}{\operatorname{argmin}} \text{ }
 \obj_\text{IOC} $}}

\end{picture}
\vspace{-0.32in}
\caption{Right: Guided cost learning uses policy optimization to adaptively sample trajectories for estimating the IOC partition function. Bottom left: PR2 learning to gently place a dish in a plate rack.
\label{fig:teaser}
\vspace{-0.20in}
}
\end{figure}

To address the challenge of representation, we propose to use expressive, nonlinear function approximators, such as neural networks, to represent the cost. This reduces the engineering burden required to deploy IOC methods, and makes it possible to learn cost functions for which expert intuition is insufficient for designing good features. Such expressive function approximators, however, can learn complex cost functions that lack the structure typically imposed by hand-designed features. To mitigate this challenge, we propose two regularization techniques for IOC, one which is general and one which is specific to episodic domains, such as to robotic manipulation skills.

In order to learn cost functions for real-world robotic tasks, our method must be able to handle unknown dynamics and high-dimensional systems. To that end, we propose a cost learning algorithm based on policy optimization with local linear models, building on prior work in reinforcement learning \cite{la-lnnpg-14}. In this approach, as illustrated in Figure~\ref{fig:teaser}, the cost function is learned in the inner loop of a policy search procedure, using samples collected for policy improvement to also update the cost function. The cost learning method itself is a nonlinear generalization of maximum entropy IOC \cite{zmbd-meirl-08}, with samples used to approximate the partition function. In contrast to previous work that optimizes the policy in the inner loop of cost learning, our approach instead updates the cost in the inner loop of policy search, making it practical and efficient. One of the benefits of this approach is that we can couple learning the cost with learning the policy for that cost. For tasks that are too complex to acquire a good global cost function from a small number of demonstrations, our method can still recover effective behaviors by running our policy learning method and retaining the learned policy. We elaborate on this further in Section~\ref{sec:generalization}.

The main contribution of our work is an algorithm that learns nonlinear cost functions from user demonstrations, at the same time as learning a policy to perform the task. Since the policy optimization ``guides'' the cost toward good regions of the space, we call this method guided cost learning. Unlike prior methods, our algorithm can handle complex, nonlinear cost function representations and high-dimensional unknown dynamics, and can be used on real physical systems with a modest number of samples. Our evaluation demonstrates the performance of our method on a set of simulated benchmark tasks, showing that it outperforms previous methods. We also evaluate our method on two real-world tasks learned directly from human demonstrations. These tasks require using torque control and vision to perform a variety of robotic manipulation behaviors, without any hand-specified cost features.

\section{Related Work}

One of the basic challenges in inverse optimal control (IOC), also known as inverse reinforcement learning (IRL), is that finding a cost or reward under which a set of demonstrations is near-optimal is underdefined. Many different costs can explain a given set of demonstrations. Prior work has tackled this issue using maximum margin formulations \cite{an-alirl-04,rbz-mmp-06}, as well as probabilistic models that explain suboptimal behavior as noise \cite{ra-birl-07,zmbd-meirl-08}. We take the latter approach in this work, building on the maximum entropy IOC model \cite{z-mpabp-10}. Although the probabilistic model mitigates some of the difficulties with IOC, there is still a great deal of ambiguity, and an important component of many prior methods is the inclusion of detailed features created using domain knowledge, which can be linearly combined into a cost, including:
indicators for successful completion of the task for a robotic ball-in-cup game \cite{bkp-reirl-11},
learning table tennis with features that include distance of the ball to the opponent's elbow \cite{mbmsp-ttirl-14},
providing the goal position as a known constraint for robotic grasping \cite{drbts-dlmioc-15},
and learning highway driving with indicators for collision and driving on the grass \cite{an-alirl-04}. While these features allow for the user to impose structure on the cost, they substantially increase the engineering burden. Several methods have proposed to learn nonlinear costs using Gaussian processes \cite{lpk-gpirl-11} and boosting \cite{rbbc-bsp-07,rsb-learch-09}, but even these methods generally operate on features rather than raw states. We instead use rich, expressive function approximators, in the form of neural networks, to learn cost functions directly on raw state representations. While neural network cost representations have previously been proposed in the literature \cite{wop-dirl-15}, they have only been applied to small, synthetic domains. Previous work has also suggested simple regularization methods for cost functions, based on minimizing $\ell_1$ or $\ell_2$ norms of the parameter vector \cite{z-mpabp-10,kprs-lofm-13} or by using unlabeled trajectories \cite{avlg-ssirl-15}. When using expressive function approximators in complex real-world tasks, we must design substantially more powerful regularization techniques to mitigate the underspecified nature of the problem, which we introduce in Section~\ref{sec:regularization}.

Another challenge in IOC is that, in order to determine the quality of a given cost function, we must solve some variant of the forward control problem to obtain the corresponding policy, and then compare this policy to the demonstrated actions.
Most early IRL algorithms required solving an MDP in the inner loop of an iterative optimization \cite{nr-airl-00,an-alirl-04,zmbd-meirl-08}.
This requires perfect knowledge of the system dynamics and access to an efficient offline solver, neither of which is available in, for instance, complex robotic control domains.
Several works have proposed to relax this requirement, for example by learning a value function instead of a cost \cite{t-lsmdp-06}, solving an approximate local control problem \cite{lk-cioc-12,ds-fat-12}, generating a discrete graph of states~\cite{bmzbf-gbioc-15,mlzlt-shgpi-15}, or only obtaining an optimal trajectory rather than policy \cite{rbz-mmp-06,rsb-learch-09}.
However, these methods still require knowledge of the system dynamics.
Given the size and complexity of the problems addressed in this work, solving the optimal control problem even approximately in the inner loop of the cost optimization is impractical. We show that good cost functions can be learned by instead learning the \emph{cost} in the inner loop of a \emph{policy optimization}. Our inverse optimal control algorithm is most closely related to other previous sample-based methods based on the principle of maximum entropy, including relative entropy IRL \cite{bkp-reirl-11} and path integral IRL \citep{kprs-lofm-13}, which can also handle unknown dynamics. However, unlike these prior methods, we adapt the sampling distribution using policy optimization. We demonstrate in our experiments that this adaptation is crucial for obtaining good results on complex robotic platforms, particularly when using complex, nonlinear cost functions.


To summarize, our proposed method is the first to combine several desirable features into a single, effective algorithm: it can handle unknown dynamics, which is crucial for real-world robotic tasks, it can deal with high-dimensional, complex systems, as in the case of real torque-controlled robotic arms, and it can learn complex, expressive cost functions, such as multilayer neural networks, which removes the requirement for meticulous hand-engineering of cost features. While some prior methods have shown good results with unknown dynamics on real robots \cite{bkp-reirl-11,kprs-lofm-13} and some have proposed using nonlinear cost functions \cite{rbz-mmp-06,rsb-learch-09,lpk-gpirl-11}, to our knowledge no prior method has been demonstrated that can provide all of these benefits in the context of complex real-world tasks.


\vspace{-.1cm}
\section{Preliminaries and Overview}

We build on the probabilistic maximum entropy inverse optimal control framework \cite{zmbd-meirl-08}. The demonstrated behavior is assumed to be the result of an expert acting stochastically and near-optimally with respect to an unknown cost function. Specifically, the model assumes that the expert samples the demonstrated trajectories $\{\traj_i\}$ from the distribution
\vspace{-.2cm}
\begin{equation}
p(\traj) = \frac{1}{Z}\exp(-\cost_\params(\traj)),\label{eq:ioc_obj}
\end{equation}
\noindent where $\traj = \{\state_1,\action_1,\dots,\state_T,\action_T\}$ is a trajectory sample, $\cost_\params(\traj) = \sum_t \cost_\params(\state_t,\action_t)$ is an unknown cost function parameterized by $\params$, and $\state_t$ and $\action_t$ are the state and action at time step $t$. Under this model, the expert is most likely to act optimally, and can generate suboptimal trajectories with a probability that decreases exponentially as the trajectories become more costly. The partition function $Z$ is difficult to compute for large or continuous domains, and presents the main computational challenge in maximum entropy IOC. The first applications of this model computed $Z$ exactly with dynamic programming \cite{zmbd-meirl-08}. However, this is only practical in small, discrete domains. More recent methods have proposed to estimate $Z$ by using the Laplace approximation \cite{lk-cioc-12}, value function approximation \cite{hk-arfdh-14}, and samples \cite{bkp-reirl-11}. As discussed in Section~\ref{sec:sampling}, we take the sample-based approach in this work, because it allows us to perform inverse optimal control without a known model of the system dynamics. This is especially important in robotic manipulation domains, where the robot might interact with a variety of objects with unknown physical properties.

To represent the cost function $\cost_\params(\st,\at)$, IOC or IRL methods typically use a linear combination of hand-crafted features, given by $\cost_\params(\at,\at) = \params\tr\mathbf{f}(\at,\st)$ \cite{an-alirl-04}. This representation is difficult to apply to more complex domains, especially when the cost must be computed from raw sensory input. In this work, we explore the use of high-dimensional, expressive function approximators for representing $\cost_\params(\st,\at)$. As we discuss in Section~\ref{sec:eval}, we use neural networks that operate directly on the robot's state, though other parameterizations could also be used with our method. Complex representations are generally considered to be poorly suited for IOC, since learning costs that associate the right element of the state with the goal of the task is already quite difficult even with simple linear representations. However, as we discuss in our evaluation, we found that such representations could be learned effectively by adaptively generating samples as part of a policy optimization procedure, as discussed in Section~\ref{sec:trajopt}.

\vspace{-.1cm}
\section{Guided Cost Learning}

In this section, we describe the guided cost learning algorithm, which combines sample-based maximum entropy IOC with forward reinforcement learning using time-varying linear models. The central idea behind this method is to adapt the sampling distribution to match the maximum entropy cost distribution $p(\traj) = \frac{1}{Z}\exp(-\cost_\params(\traj))$, by directly optimizing a trajectory distribution with respect to the current cost $\cost_\params(\traj)$ using a sample-efficient reinforcement learning algorithm. Samples generated on the physical system are used both to improve the policy and more accurately estimate the partition function $Z$. In this way, the reinforcement learning step acts to ``guide'' the sampling distribution toward regions where the samples are more useful for estimating the partition function. We will first describe how the IOC objective in~\eqref{eq:ioc_obj} can be estimated with samples, and then describe how reinforcement learning can adapt the sampling distribution.

\subsection{Sample-Based Inverse Optimal Control}
\label{sec:sampling}


In the sample-based approach to maximum entropy IOC, the partition function $Z = \int \exp(-\cost_\params(\traj)) d\traj$ is estimated with samples from a background distribution $q(\traj)$. Prior sample-based IOC methods use a linear representation of the cost function, which simplifies the corresponding cost learning problem \cite{bkp-reirl-11,kprs-lofm-13}. In this section, we instead derive a sample-based approximation for the IOC objective for a general nonlinear parameterization of the cost function. The negative log-likelihood corresponding to the IOC model in~\eqref{eq:ioc_obj} is given by:
\begin{align*}
\obj_\text{IOC}(\params) &= \frac{1}{N}\!\sum_{\traj_i \in \mathcal{D}_\text{demo}} \!\! \cost_\params(\traj_i) + \log Z\\
 &\approx\!\frac{1}{N}\!\sum_{\traj_i \in \mathcal{D}_\text{demo}} \!\!\!\!\cost_\params(\traj_i) +\!\log \frac{1}{M}\!\sum_{\traj_j \in \mathcal{D}_\text{samp}}\!\!\!\frac{\exp(-\cost_\params(\traj_j))}{q(\traj_j)},
\end{align*}
\noindent where $\mathcal{D}_\text{demo}$ denotes the set of $N$ demonstrated trajectories, $\mathcal{D}_\text{samp}$ the set of $M$ background samples, and $q$ denotes the background distribution from which trajectories $\traj_j$ were sampled. Prior methods have chosen $q$ to be uniform \cite{bkp-reirl-11} or to lie in the vicinity of the demonstrations~\cite{kprs-lofm-13}.
To compute the gradients of this objective with respect to the cost parameters $\params$, let \mbox{$w_j = \frac{\exp(-\cost_\params(\traj_j))}{q(\traj_j)}$} and $Z = \sum_j w_j$. The gradient is then given by:
\begin{align}
&\frac{d\obj_\text{IOC}}{d\params} =\nonumber
\frac{1}{N}\sum_{\traj_i \in \mathcal{D}_\text{demo}} \frac{d \cost_\params}{d \params}(\traj_i) - \frac{1}{Z} \sum_{\traj_j \in \mathcal{D}_\text{samp}} w_j \frac{d \cost_\params}{d \params}(\traj_j)
\end{align}
When the cost is represented by a neural network or some other function approximator, this gradient can be computed efficiently by backpropagating $- \frac{w_j}{Z}$ for each trajectory $\traj_j \in \mathcal{D}_\text{samp}$ and $\frac{1}{N}$ for each trajectory $\traj_i \in \mathcal{D}_\text{demo}$. 


\subsection{Adaptive Sampling via Policy Optimization}
\label{sec:trajopt}

The choice of background sample distribution $q(\traj)$ for estimating the objective $\obj_\text{IOC}$ is critical for successfully applying the sample-based IOC algorithm. The optimal importance sampling distribution for estimating the partition function $\int \exp(-\cost_\params(\tau)) d\tau$ is $q(\tau) \propto |\exp(-\cost_\params(\tau))| = \exp(-\cost_\params(\tau))$.
Designing a single background distribution $q(\traj)$ is therefore quite difficult when the cost $\cost_\params$ is unknown. Instead, we can adaptively refine $q(\traj)$ to generate more samples in those regions of the trajectory space that are good according to the current cost function $\cost_\params(\traj)$. To this end, we interleave the IOC optimization, which attempts to find the cost function that maximizes the likelihood of the demonstrations, with a policy optimization procedure, which improves the trajectory distribution $q(\traj)$ with respect to the current cost.

Since one of the main advantages of the sample-based IOC approach is the ability to handle unknown dynamics, we must also choose a policy optimization procedure that can handle unknown dynamics. To this end, we adapt the method presented by \citet{la-lnnpg-14}, which performs policy optimization under unknown dynamics by iteratively fitting time-varying linear dynamics to samples from the current trajectory distribution $q(\traj)$, updating the trajectory distribution using a modified LQR backward pass, and generating more samples for the next iteration. The trajectory distributions generated by this method are Gaussian, and each iteration of the policy optimization procedure satisfies a KL-divergence constraint of the form \mbox{$\kl(q(\traj)\|\hat{q}(\traj)) \leq \epsilon$}, which prevents the policy from changing too rapidly~\cite{bagnell2003covariant,pma-reps-10,rawlik2013stochastic}.
This has the additional benefit of not overfitting to poor initial estimates of the cost function. With a small modification, we can use this algorithm to optimize a maximum entropy version of the objective, given by $\min_q E_q[\cost_\params(\traj)] - \ent(\traj)$, as discussed in prior work \cite{la-lnnpg-14}. This variant of the algorithm allows us to recover the trajectory distribution $q(\traj) \propto \exp(-\cost_\params(\traj))$ at convergence \cite{z-mpabp-10}, a good distribution for sampling. For completeness, this policy optimization procedure is summarized in Appendix~\ref{app:trajopt}.

\begin{algorithm}[t]
{\small
\caption{Guided cost learning}
\label{alg:adaptive}
\begin{algorithmic}[1]
\STATE Initialize $q_k(\traj)$ as either a random initial controller or from demonstrations
\FOR{iteration $i=1$ to $I$}
\STATE Generate samples $\mathcal{D}_\text{traj}$ from $q_k(\traj)$
\STATE Append samples: $\mathcal{D}_\text{samp} \leftarrow \mathcal{D}_\text{samp} \cup \mathcal{D}_\text{traj}$
\STATE Use $\mathcal{D}_\text{samp}$ to update cost $\cost_\params$ using Algorithm~\ref{alg:batch_ioc}
\STATE Update $q_k(\traj)$ using $\mathcal{D}_\text{traj}$ and the method from \cite{la-lnnpg-14} to obtain $q_{k+1}(\traj)$
\ENDFOR
\STATE {\bf return} optimized cost parameters $\params$ and trajectory distribution $q(\traj)$
\end{algorithmic}
}
\end{algorithm}

Our sample-based IOC algorithm with adaptive sampling is summarized in Algorithm~\ref{alg:adaptive}. We call this method guided cost learning because policy optimization is used to guide sampling toward regions with lower cost. The algorithm consists of taking successive policy optimization steps, each of which generates samples $\mathcal{D}_\text{traj}$ from the latest trajectory distribution $q_k(\traj)$. After sampling, the cost function is updated using all samples collected thus far for the purpose of policy optimization. No additional background samples are required for this method. This procedure returns both a learned cost function $\cost_\params(\st,\at)$ and a trajectory distribution $q(\traj)$, which corresponds to a time-varying linear-Gaussian controller $q(\at|\st)$. This controller can be used to execute the learned behavior.

\vspace{-.1cm}
\subsection{Cost Optimization and Importance Weights}
\begin{algorithm}[b]
{\small
\caption{Nonlinear IOC with stochastic gradients}
\label{alg:batch_ioc}
\begin{algorithmic}[1]
\FOR{iteration $k=1$ to $K$}
\STATE Sample demonstration batch $\hat{\mathcal{D}}_\text{demo} \subset \mathcal{D}_\text{demo}$
\STATE Sample background batch $\hat{\mathcal{D}}_\text{samp} \subset \mathcal{D}_\text{samp}$
\STATE Append demonstration batch to background batch: \mbox{$\hat{\mathcal{D}}_\text{samp} \leftarrow \hat{\mathcal{D}}_\text{demo} \cup \hat{\mathcal{D}}_\text{samp}$}
\STATE Estimate $\frac{d\obj_\text{IOC}}{d\params}(\params)$ using $\hat{\mathcal{D}}_\text{demo}$ and $\hat{\mathcal{D}}_\text{samp}$
\STATE Update parameters $\params$ using gradient $\frac{d\obj_\text{IOC}}{d\params}(\params)$
\ENDFOR
\STATE {\bf return} optimized cost parameters $\params$
\end{algorithmic}
}
\end{algorithm}

The IOC objective can be optimized using standard nonlinear optimization methods and the gradient $\frac{d\obj_\text{IOC}}{d\params}$. Stochastic gradient methods are often preferred for high-dimensional function approximators, such as the neural networks. Such methods are straightforward to apply to objectives that factorize over the training samples, but the partition function does not factorize trivially in this way. Nonetheless, we found that our objective could still be optimized with stochastic gradient methods by sampling a subset of the demonstrations and background samples at each iteration. When the number of samples in the batch is small, we found it necessary to add the sampled demonstrations to the background sample set as well; without adding the demonstrations to the sample set, the objective can become unbounded and frequently does in practice.
The stochastic optimization procedure is presented in Algorithm~\ref{alg:batch_ioc}, and is straightforward to implement with most neural network libraries based on backpropagation.

Estimating the partition function requires us to use importance sampling. Although prior work has suggested dropping the importance weights \cite{kprs-lofm-13,ab-meirlpi-11}, we show in Appendix~\ref{app:consistent} that this produces an \mbox{inconsistent} likelihood estimate and fails to recover good cost functions. Since our samples are drawn from multiple distributions, we compute a fusion distribution to evaluate the importance weights.
Specifically, if we have samples from $k$ distributions $q_1(\traj),\dots,q_\kappa(\traj)$, we can construct a consistent estimator of the expectation of a function $f(\traj)$ under a uniform distribution as \mbox{$E[f(\traj)] \approx \frac{1}{M} \sum_{\traj_j}  \frac{1}{\frac{1}{k} \sum_\kappa q_\kappa(\traj_j)} f(\traj_j)$}.
Accordingly, the importance weights are \mbox{$z_j = [\frac{1}{k} \sum_\kappa q_\kappa(\traj_j)]^{-1}$}, and the objective is now:
\vspace{-.17cm}
\begin{align*}
\obj_\text{IOC}(\params) =&
\frac{1}{N}\!\!\sum_{\traj_i \in \mathcal{D}_\text{demo}} \!\! \cost_\params(\traj_i)
+ \log \frac{1}{M}\!\! \sum_{\traj_j \in \mathcal{D}_\text{samp}}  \!\!\!\! z_j \exp(-\cost_\params(\traj_j))
\vspace{-.56cm}
\end{align*}
The distributions $q_\kappa$ underlying background samples are obtained from the controller at iteration $k$. We must also append the demonstrations to the samples in Algorithm~\ref{alg:batch_ioc}, yet the distribution that generated the demonstrations is unknown. To estimate it, we assume the demonstrations come from a single Gaussian trajectory distribution and compute its empirical mean and variance. We found this approximation sufficiently accurate for estimating the importance weights of the demonstrations, as shown in Appendix~\ref{app:consistent}.

\vspace{-.14cm}
\subsection{Learning Costs and Controllers}
\label{sec:generalization}

In contrast to many previous IOC and IRL methods, our approach can be used to learn a cost while simultaneously optimizing the policy for a new instance of the task not in the demos, such as a new position of a target cup for a pouring task, as shown in our experiments. Since the algorithm produces both a cost function $\cost_\params(\st,\at)$ and a controller $q(\at|\st)$ that optimizes this cost on the new task instance, we can directly use this controller to execute the desired behavior. In this way, the method actually learns a policy from demonstration, using the additional knowledge that the demonstrations are near-optimal under some unknown cost function, similar to recent work on IOC by direct loss minimization \cite{drbts-dlmioc-15}. The learned cost function $\cost_\params(\st,\at)$ can often also be used to optimize new policies for new instances of the task without additional cost learning. However, we found that on the most challenging tasks we tested, running policy learning with IOC in the loop for each new task instance typically succeeded more frequently than running IOC once and reusing the learned cost. We hypothesize that this is because training the policy on a new instance of the task provides the algorithm with additional information about task variation, thus producing a better cost function and reducing overfitting. The intuition behind this hypothesis is that the demonstrations only cover a small portion of the degrees of variation in the task. Observing samples from a new task instance provides the algorithm with a better idea of the particular factors that distinguish successful task executions from failures.





\vspace{-.2cm}
\section{Representation and Regularization}
\label{sec:regularization}

We parametrize our cost functions as neural networks, expanding their expressive power and enabling IOC to be applied to the state of a robotic system directly, without hand-designed features.
Our experiments in Section~\ref{sec:real} confirm that an affine cost function is not expressive enough to learn some behaviors.
Neural network parametrizations are particularly useful for learning visual representations on raw image pixels.
In our experiments, we make use of the unsupervised visual feature learning method developed by~\citet{ftddla-dsae-15} to learn cost functions that depend on visual input.
Learning cost functions on raw pixels is an interesting direction for future work, which we discuss in Section~\ref{sec:discussion}.


While the expressive power of nonlinear cost functions provide a range of benefits, they introduce significant model complexity to an already underspecified IOC objective.
To mitigate this challenge, we propose two regularization methods for IOC.
Prior methods regularize the IOC objective by penalizing the $\ell_1$ or $\ell_2$ norm of the cost parameters $\params$~\cite{z-mpabp-10,kprs-lofm-13}.
For high-dimensional nonlinear cost functions, this regularizer is often insufficient, since different entries in the parameter vector can have drastically different effects on the cost.
We use two regularization terms. The first term encourages the cost of demo and sample trajectories to change locally at a constant rate (lcr), by penalizing the second time derivative:
\vspace{-.08cm}
\begin{equation*}
\smoothreg(\traj) \!= \!\!\sum_{x_t \in \traj} [(\cost_\params(x_{t+1})-\cost_\params(x_t)) - (\cost_\params(x_t)-\cost_\params(x_{t-1}))]^2
\vspace{-.1cm}
\end{equation*}
This term reduces high-frequency variation that is often symptomatic of overfitting, making the cost easier to reoptimize. Although sharp changes in the cost slope are sometimes preferred, we found that temporally slow-changing costs were able to adequately capture all of the behaviors in our experiments.

The second regularizer is more specific to one-shot episodic tasks, and it encourages the cost of the states of a demo trajectory to decrease strictly monotonically in time using a squared hinge loss:
\vspace{-.08cm}
\begin{equation*}
\monoreg(\traj) = \sum_{x_t \in \traj} [\max(0,\cost_\params(x_t)-\cost_\params(x_{t-1})-1)]^2
\vspace{-.1cm}
\end{equation*}
The rationale behind this regularizer is that, for tasks that essentially consist of reaching a target condition or state, the demonstrations typically make monotonic progress toward the goal on some (potentially nonlinear) manifold. While this assumption does not always hold perfectly, we again found that this type of regularizer improved performance on the tasks in our evaluation. We show a detailed comparison with regard to both regularizers in Appendix~\ref{app:ablation}.


\vspace{-.2cm}

\section{Experimental Evaluation}
\label{sec:eval}

We evaluated our sampling-based IOC algorithm on a set of robotic control tasks, both in simulation and on a real robotic platform. Each of the experiments involve complex second order dynamics with force or torque control and no manually designed cost function features, with the raw state provided as input to the learned cost function.

We also tested the consistency of our algorithm on a toy point mass example for which the ground truth distribution is known. These experiments, discussed fully in Appendix~\ref{app:consistent}, show that using a maximum entropy version of the policy optimization objective (see Section~\ref{sec:trajopt}) and using importance weights are both necessary for recovering the true distribution.

\vspace{-.1cm}
\subsection{Simulated Comparisons}
\label{sec:sim}

In this section, we provide simulated comparisons between guided cost learning and prior sample-based methods. We focus on task performance and sample complexity, and also perform comparisons across two different sampling distribution initializations and regularizations (in Appendix~\ref{app:ablation}).

\begin{figure}
\setlength{\unitlength}{0.5\columnwidth}
\begin{picture}(1.0,2.33) \linethickness{0.5pt}

\put(0.0,1.56){\includegraphics[height=0.39\columnwidth]{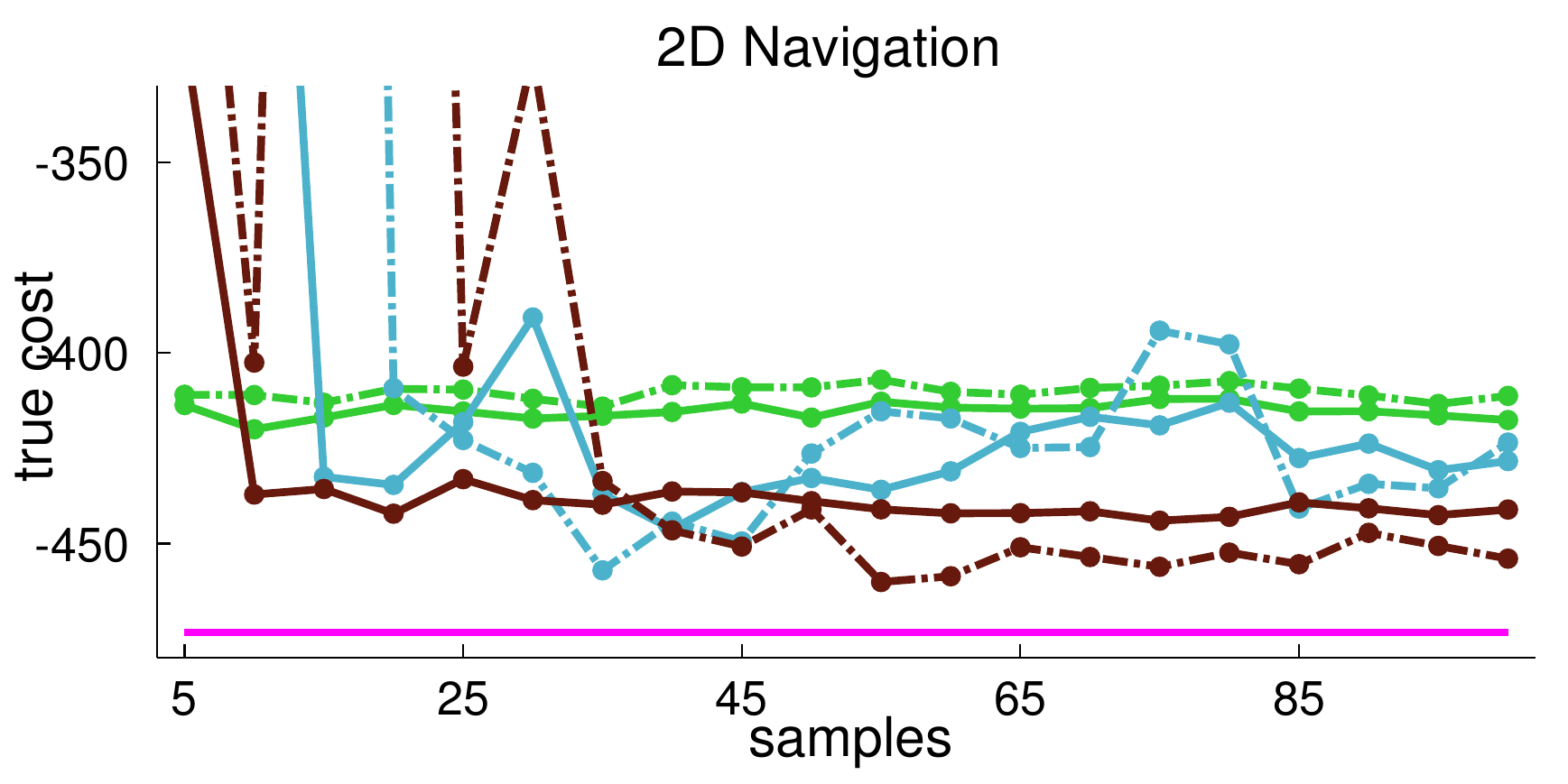}}
\put(1.52,1.99){\includegraphics[height=0.18\columnwidth]{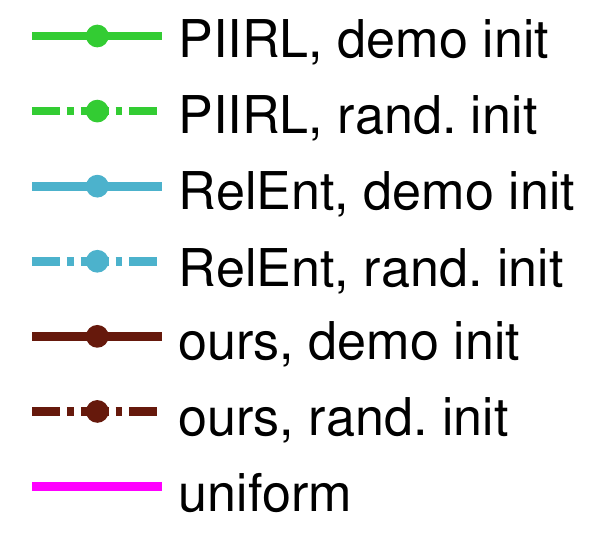}}

\put(1.53,1.55){\includegraphics[height=0.22\columnwidth]{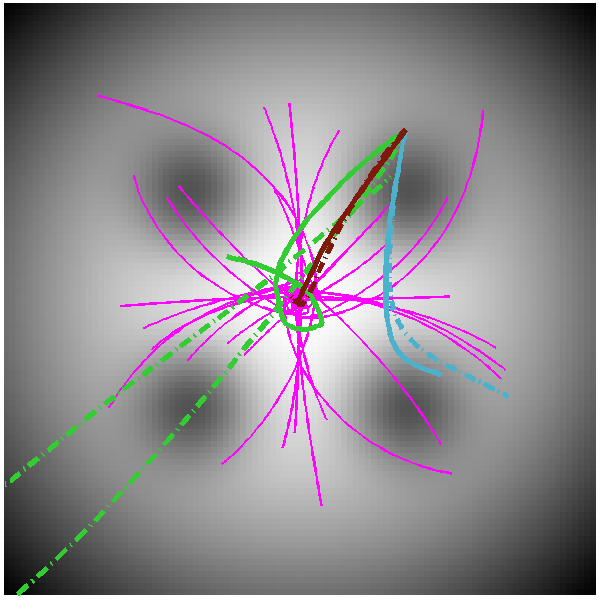}}

\put(1.53,1.0){\includegraphics[height=0.24\columnwidth]{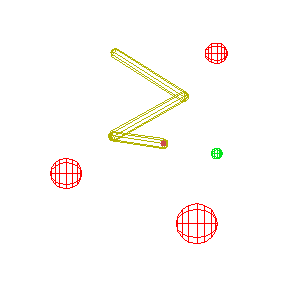}}
\put(1.56,0.99){\small{green: goal}}
\put(1.56,0.91){\small{red: obstacles}}

\put(0.0,0.78){\includegraphics[height=0.39\columnwidth]{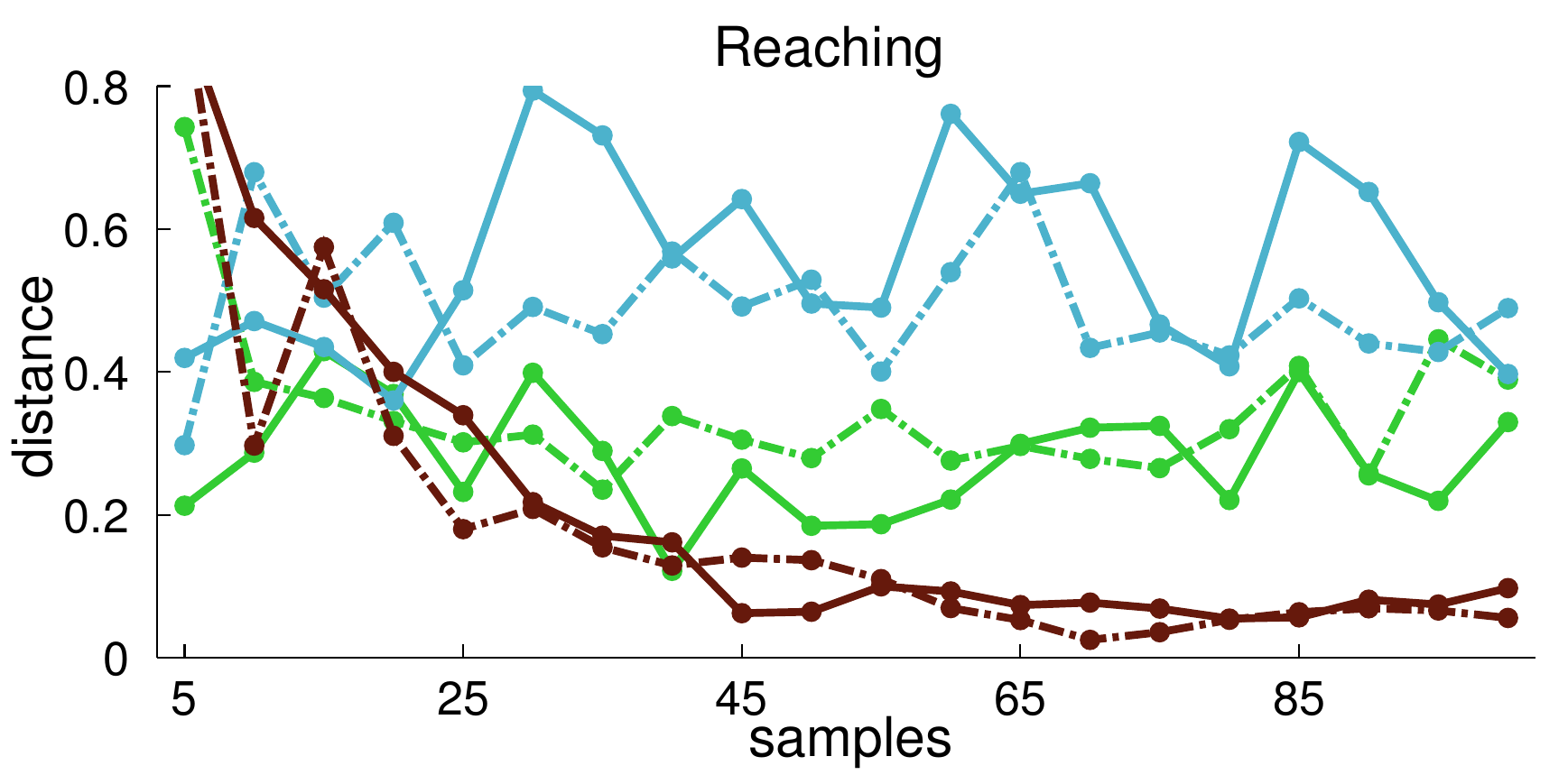}}

\put(1.5,0.30){\includegraphics[height=0.23\columnwidth]{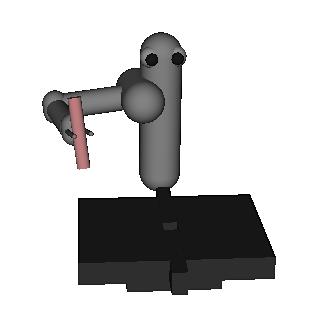}}
\put(1.5,-0.05){\includegraphics[height=0.23\columnwidth]{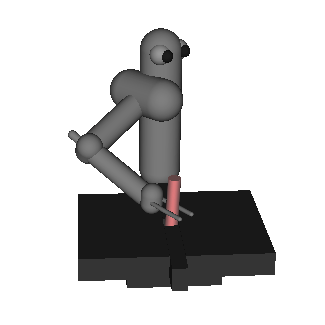}}
\put(1.8,0.70){\footnotesize{initial}}
\put(1.84,0.63){\footnotesize{state}}
\put(1.83,0.32){\footnotesize{goal}}
\put(1.83,0.25){\footnotesize{state}}

\put(0.0,0.0){\includegraphics[height=0.39\columnwidth]{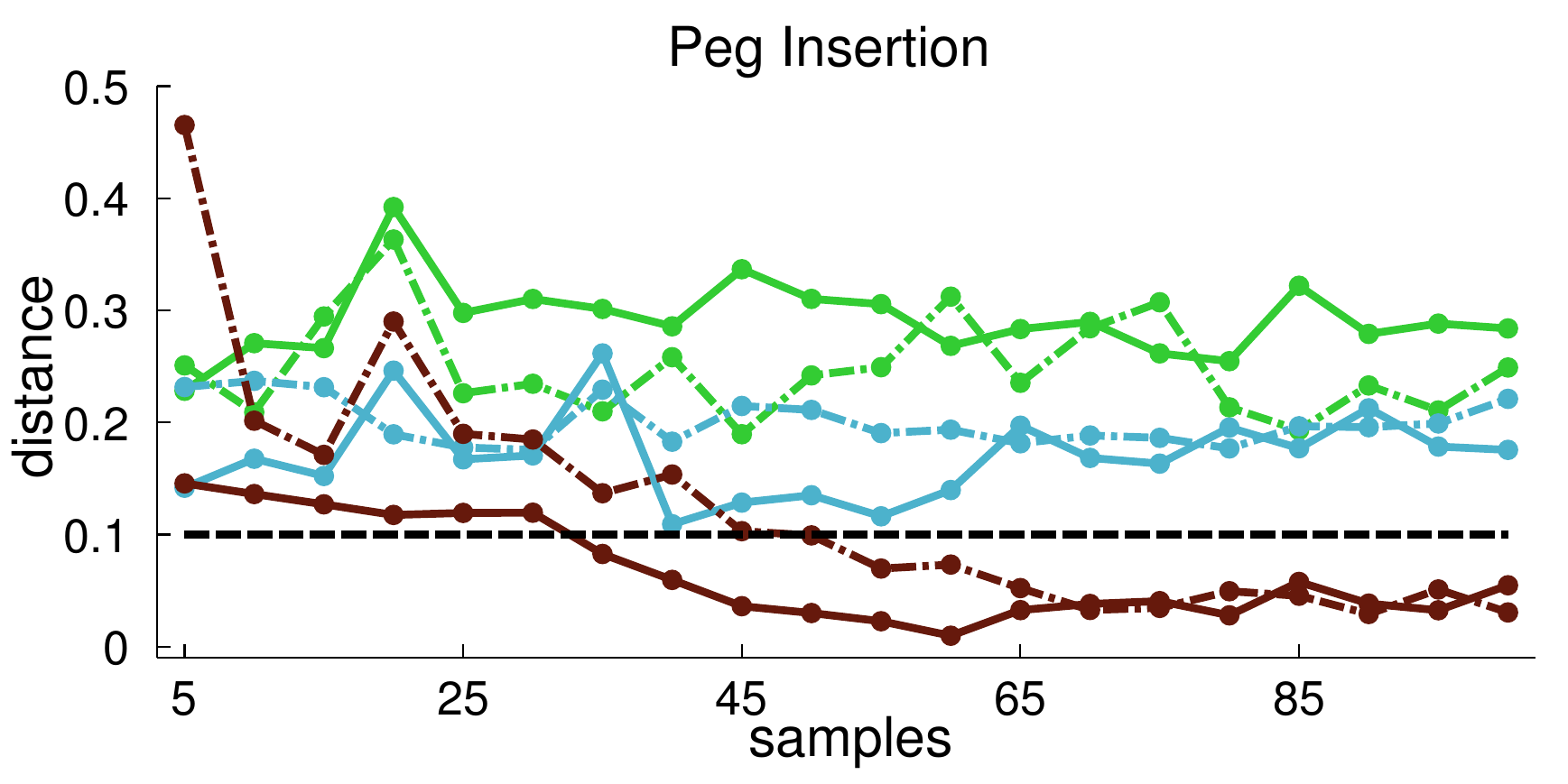}}

\end{picture}
\vspace{-0.15in}
\caption{Comparison to prior work on simulated 2D navigation, reaching, and peg insertion tasks. Reported performance is averaged over 4 runs of IOC on 4 different initial conditions . For peg insertion, the depth of the hole is 0.1m, marked as a dashed line. Distances larger than this amount failed to insert the peg.
\label{fig:2d}
\vspace{-0.2in}
}
\end{figure}
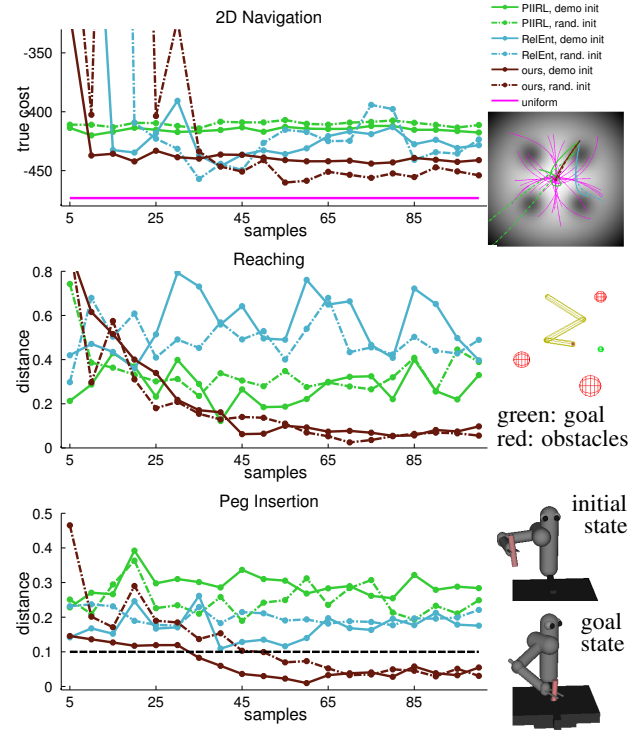



To compare guided cost learning to prior methods, we ran experiments on three simulated tasks of varying difficulty, all using the MuJoCo physics simulator~\cite{tet-mjc-12}. The first task is 2D navigation around obstacles, modeled on the task by~\citet{lk-cioc-12}. This task has simple, linear dynamics and a low-dimensional state space, but a complex cost function, which we visualize in Figure~\ref{fig:2d}. The second task involves a 3-link arm reaching towards a goal location in 2D, in the presence of physical obstacles. The third, most challenging, task is 3D peg insertion with a 7 DOF arm. This task is significantly more difficult than tasks evaluated in prior IOC work as it involves complex contact dynamics between the peg and the table and high-dimensional, continuous state and action spaces. The arm is controlled by selecting torques at the joint motors at 20 Hz. More details on the experimental setup are provided in Appendix~\ref{app:tasks}.

In addition to the expert demonstrations, prior methods require a set of ``suboptimal'' samples for estimating the partition function. We obtain these samples in one of two ways: by using a baseline random controller that randomly explores around the initial state (random), and by fitting a linear-Gaussian controller to the demonstrations (demo). The latter initialization typically produces a motion that tracks the average demonstration with variance proportional to the variation between demonstrated motions.

Between 20 and 32 demonstrations were generated from a policy learned using the method of \citet{la-lnnpg-14}, with a ground truth cost function determined by the agent's pose relative to the goal. We found that for the more precise peg insertion task, a relatively complex ground truth cost function was needed to afford the necessary degree of precision. We used a cost function of the form $w d^2 + v \log(d^2 + \alpha)$, where $d$ is the distance between the two tips of the peg and their target positions, and $v$ and $\alpha$ are constants. Note that the affine cost is incapable of exactly representing this function. We generated demonstration trajectories under several different starting conditions. For 2D navigation, we varied the initial position of the agent, and for peg insertion, we varied the position of the peg hole. We then evaluated the performance of our method and prior sample-based methods~\cite{kprs-lofm-13,bkp-reirl-11} on each task from four arbitrarily-chosen test states. We chose these prior methods because, to our knowledge, they are the only methods which can handle unknown dynamics.

We used a neural network cost function with two hidden layers with 24--52 units and rectifying nonlinearities of the form $\max(z,0)$ followed by linear connections to a set of features $\mathbf{y}_t$, which had a size of 20 for the 2D navigation task and 100 for the other two tasks.  The cost is then given by
\vspace{-0.3cm}
\begin{equation}
\cost_\params(\st,\at) = \| A\mathbf{y}_t + b \|^2 + w_\action \| \at \|^2
\label{eqn:costnneq}
\vspace{-0.3cm}
\end{equation}

with a fixed torque weight $w_\action$ and the parameters consisting of $A$, $b$, and the network weights. These cost functions range from about 3,000 parameters for the 2D navigation task to 16,000 parameters for peg insertion. For further details, see Appendix~\ref{app:nn}. Although the prior methods learn only linear cost functions, we can extend them to the nonlinear setting following the derivation in Section~\ref{sec:sampling}.

Figure~\ref{fig:2d} illustrates the tasks and shows results for each method after different numbers of samples from the test condition. In our method, five samples were used at each iteration of policy optimization, while for the prior methods, the number of samples corresponds to the number of ``suboptimal'' samples provided for cost learning. For the prior methods, additional samples were used to optimize the learned cost. The results indicate that our method is generally capable of learning tasks that are more complex than the prior methods, and is able to effectively handle complex, high-dimensional neural network cost functions. In particular, adding more samples for the prior methods generally does not improve their performance, because all of the samples are drawn from the same distribution.

\subsection{Real-World Robotic Control}
\label{sec:real}

We also evaluated our method on a set of real robotic manipulation tasks using the PR2 robot, with comparisons to relative entropy IRL, which we found to be the better of the two prior methods in our simulated experiments. We chose two robotic manipulation tasks which involve complex dynamics and interactions with delicate objects, for which it is challenging to write down a cost function by hand.
For all methods, we used a two-layer neural network cost parametrization and the regularization objective described in Section~\ref{sec:regularization}, and compared to an affine cost function on one task to evaluate the importance of non-linear cost representations. The affine cost followed the form of equation~\ref{eqn:costnneq} but with $\mathbf{y}_t$ equal to the input $\mathbf{x}_t$.\footnote{Note that a cost function that is quadratic in the state is linear in the coefficients of the monomials, and therefore corresponds to a linear parameterization.} For both tasks, between 25 and 30 human demonstrations were provided via kinesthetic teaching, and each IOC algorithm was initialized by automatically fitting a controller to the demonstrations that roughly tracked the average trajectory. Full details on both tasks are in Appendix~\ref{app:tasks}, and summaries are below.

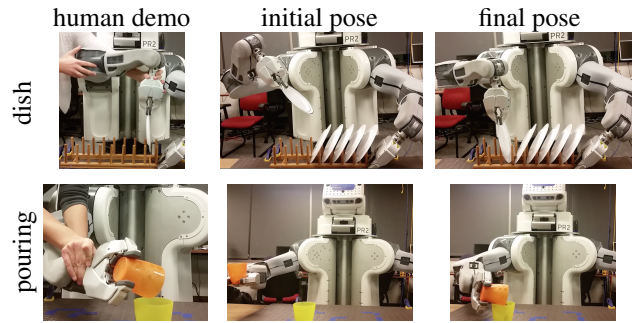
\begin{figure}
\vspace{-.2cm}
\setlength{\unitlength}{0.5\columnwidth}
\begin{picture}(1.99,1.0) \linethickness{0.5pt}

\put(0.15,0.44){\includegraphics[height=0.22\columnwidth]{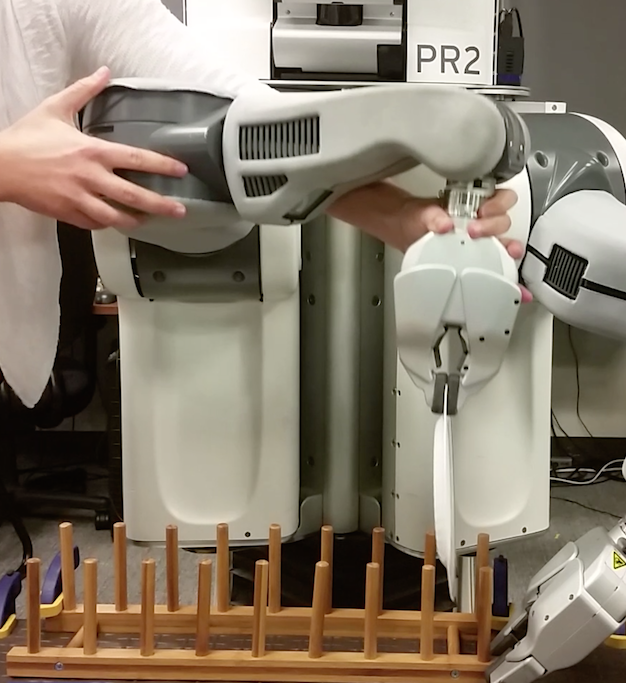}}
\put(0.67,0.44){\includegraphics[height=0.22\columnwidth]{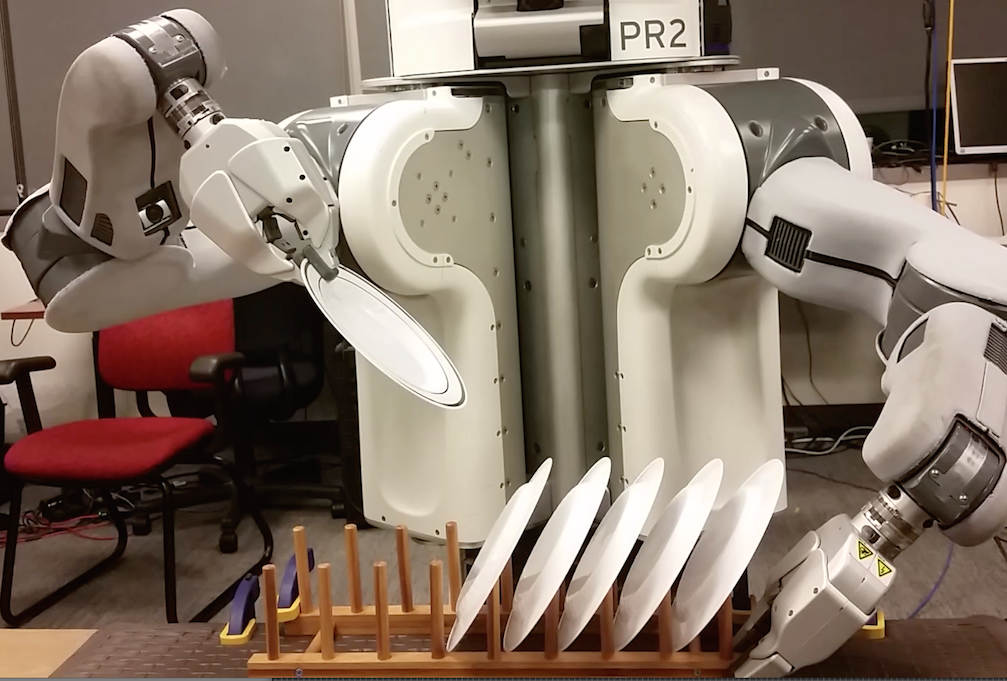}}
\put(1.36,0.44){\includegraphics[height=0.22\columnwidth]{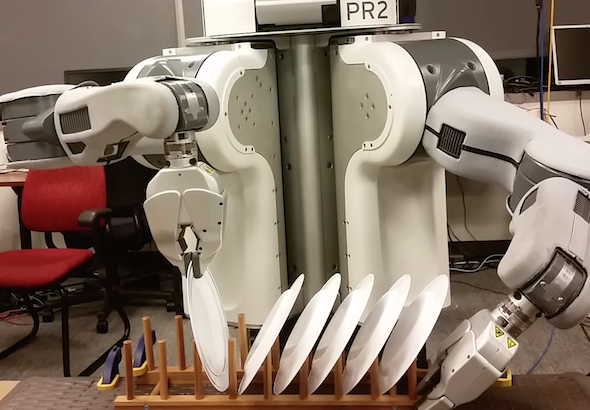}}
\put(0.13,0.9){human demo}
\put(0.8,0.9){initial pose}
\put(1.5,0.9){final pose}

\put(0.1,-0.05){\includegraphics[height=0.22\columnwidth]{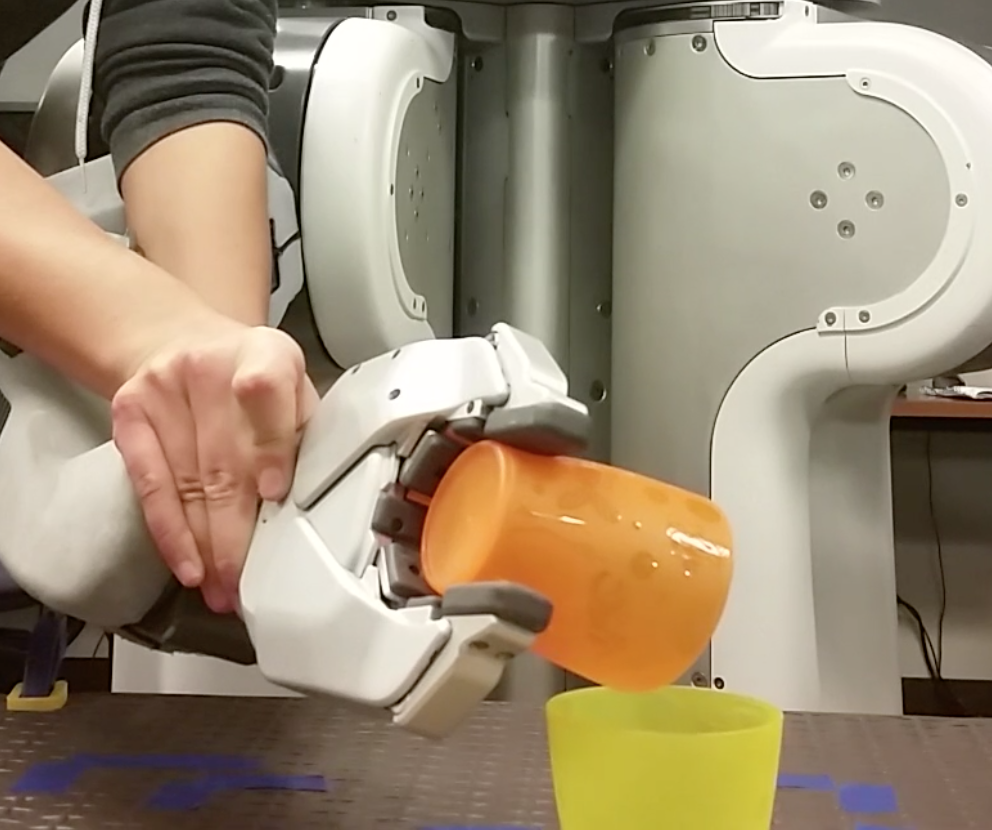}}
\put(0.69,-0.05){\includegraphics[height=0.22\columnwidth]{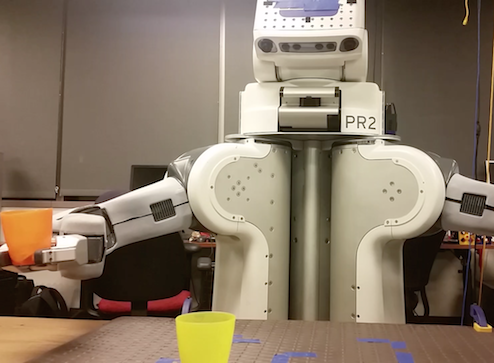}}
\put(1.41,-0.05){\includegraphics[height=0.22\columnwidth]{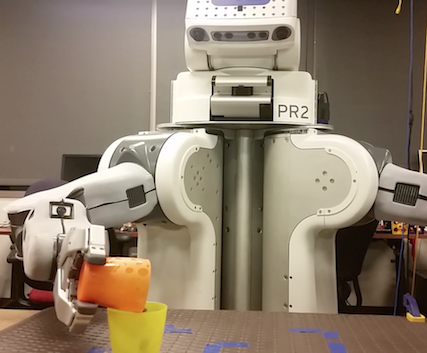}}
\put(0.0,0.038){\rotatebox{90}{pouring}}
\put(0.0,0.58){\rotatebox{90}{dish}}

\end{picture}
\vspace{-0.22in}
\caption{Dish placement and pouring tasks. The robot learned to place the plate gently into the correct slot, and to pour almonds, localizing the target cup using unsupervised visual features. A video of the learned controllers can be found at \mbox{\url{http://rll.berkeley.edu/gcl}}
\label{fig:tasks}
\vspace{-0.22in}
}
\end{figure}

In the first task, illustrated in Figure~\ref{fig:tasks}, the robot must gently place a grasped plate into a specific slot of dish rack. The state space consists of the joint angles, the pose of the gripper relative to the target pose, and the time derivatives of each; the actions correspond to torques on the robot's motors; and the input to the cost function is the pose and velocity of the gripper relative to the target position. Note that we do not provide the robot with an existing trajectory tracking controller or any manually-designed policy representation beyond linear-Gaussian controllers, in contrast to prior methods that use trajectory following \cite{kprs-lofm-13} or dynamic movement primitives with features \cite{bkp-reirl-11}. Our attempt to design a hand-crafted cost function for inserting the plate into the dish rack produced a fast but overly aggressive behavior that cracked one of the plates during learning.

The second task, also shown in Figure~\ref{fig:tasks}, consisted of pouring almonds from one cup to another. In order to succeed, the robot must keep the cup upright until reaching the target cup, then rotate the cup so that the almonds are poured. Instead of including the position of the target cup in the state space, we train autoencoder features from camera images captured from the demonstrations and add a pruned feature point representation and its time derivative to the state, as proposed by~\citet{ftddla-dsae-15}. The input to the cost function includes these visual features, as well as the pose and velocity of the gripper. Note that the position of the target cup can only be obtained from the visual features, so the algorithm must learn to use them in the cost function in order to succeed at the task.

\vspace{-0.1cm}
\begin{table}[]
{\footnotesize
  \begin{center}
    \begin{tabular}{| l || c || c | c | c |}
        \hline
    \emph{dish} (NN) &\!RelEnt IRL\!&\!GCL $q(\at|\st)$\!\!& GCL reopt. \\
    \hline
    success rate & 0\% & \textbf{100\%} & \textbf{100\%} \\
    \hline
    \# samples &  100 & 90 & 90\\
    \hline
    \hline
    \emph{pouring} (NN) &\!RelEnt IRL\!&\!\!GCL $q(\at|\st)$\!\!&\!\!GCL reopt.\!\!\\
    \hline
    success rate & 10\% & \textbf{84.7\%}  & 34\%  \\
    \hline
    \# samples &  150,150 & 75,130 & 75,130\\
    \hline
     \hline
    \emph{pouring} (affine)\!\!&\!RelEnt IRL\!&\!\!GCL $q(\at|\st)$\!\!& GCL reopt. \\
    \hline
    success rate & 0\% & 0\% & -- \\
    \hline
    \# samples &  150 & 120 & -- \\
    \hline    
    \end{tabular}
  \end{center}
}
\vspace{-0.1in}
\caption{Performance of guided cost learning (GCL) and relative entropy (RelEnt) IRL on placing a dish into a rack and pouring almonds into a cup.
Sample counts are for IOC, omitting those for optimizing the learned cost.
An affine cost is insufficient for representing the pouring task, thus motivating using a neural network cost (NN).
The pouring task with a neural network cost is evaluated for two positions of the target cup; average performance is reported. 
  }
  \label{tbl:realresults}
\vspace{-0.2in}
\end{table}

The results, presented in Table~\ref{tbl:realresults}, show that our algorithm successfully learned both tasks. The prior relative entropy IRL algorithm could not acquire a suitable cost function, due to the complexity of this domain. On the pouring task, where we also evaluated a simpler affine cost function, we found that only the neural network representation could recover a successful behavior, illustrating the need for rich and expressive function approximators when learning cost functions directly on raw state representations.\footnote{We did attempt to learn costs directly on image pixels, but found that the problem was too underdetermined to succeed. Better image-specific regularization is likely required for this.}

\vspace{-0.1cm}
The results in Table~\ref{tbl:realresults} also evaluate the generalizability of the cost function learned by our method and prior work. On the dish rack task, we can use the learned cost to optimize new policies for different target dish positions successfully, while the prior method does not produce a generalizable cost function. On the harder pouring task, we found that the learned cost succeeded less often on new positions. However, as discussed in Section~\ref{sec:generalization}, our method produces both a policy $q(\at|\st)$ and a cost function $\cost_\params$ when trained on a novel instance of the task, and although the learned cost functions for this task were worse, the learned policy succeeded on the test positions when optimized with IOC in the inner loop using our algorithm. This indicates an interesting property of our approach: although the learned cost function is local in nature due to the choice of sampling distribution, the learned policy tends to succeed even when the cost function is too local to produce good results in very different situations. An interesting avenue for future work is to further explore the implications of this property, and to improve the generalizability of the learned cost by successively training policies on different novel instances of the task until enough global training data is available to produce a cost function that is a good fit to the demonstrations in previously unseen parts of the state space.

\section{Discussion and Future Work}
\label{sec:discussion}

We presented an inverse optimal control algorithm that can learn complex, nonlinear cost representations, such as neural networks, and can be applied to high-dimensional systems with unknown dynamics. Our algorithm uses a sample-based approximation of the maximum entropy IOC objective, with samples generated from a policy learning algorithm based on local linear models \cite{la-lnnpg-14}. To our knowledge, this approach is the first to combine the benefits of sample-based IOC under unknown dynamics with nonlinear cost representations that directly use the raw state of the system, without the need for manual feature engineering. This allows us to apply our method to a variety of real-world robotic manipulation tasks. Our evaluation demonstrates that our method outperforms prior IOC algorithms on a set of simulated benchmarks, and achieves good results on several real-world tasks.



Our evaluation shows that our approach can learn good cost functions for a variety of simulated tasks.
For complex robotic motion skills, the learned cost functions tend to explain the demonstrations only locally. This makes them difficult to reoptimize from scratch for new conditions. It should be noted that this challenge is not unique to our method. In our comparisons, no prior sample-based method was able to learn good global costs for these tasks. However, since our method interleaves cost optimization with policy learning, it still recovers successful policies for these tasks. For this reason, we can still learn from demonstration simply by retaining the learned policy, and discarding the cost function. This allows us to tackle substantially more challenging tasks that involve direct torque control of real robotic systems with feedback from vision.

To incorporate vision into our experiments, we used unsupervised learning to acquire a vision-based state representation, following prior work \cite{ftddla-dsae-15}. An exciting avenue for future work is to extend our approach to learn cost functions directly from natural images. The principal challenge for this extension is to avoid overfitting when using substantially larger and more expressive networks. Our current regularization techniques mitigate overfitting to a high degree, but visual inputs tend to vary dramatically between demonstrations and on-policy samples, particularly when the demonstrations are provided by a human via kinesthetic teaching. One promising avenue for mitigating these challenges is to introduce regularization methods developed for domain adaptation in computer vision \cite{thds-dtad-15}, to encode the prior knowledge that demonstrations have similar visual features to samples.


\section*{Acknowledgements}

This research was funded in part by ONR through a Young Investigator Program
award, the Army Research Office through the MAST program, and an NSF
fellowship. We thank Anca Dragan for thoughtful discussions.


\bibliography{references}

\begin{thebibliography}{31}
\providecommand{\natexlab}[1]{#1}
\providecommand{\url}[1]{\texttt{#1}}
\expandafter\ifx\csname urlstyle\endcsname\relax
  \providecommand{\doi}[1]{doi: #1}\else
  \providecommand{\doi}{doi: \begingroup \urlstyle{rm}\Url}\fi

\bibitem[Abbeel \& Ng(2004)Abbeel and Ng]{an-alirl-04}
Abbeel, P. and Ng, A.
\newblock Apprenticeship learning via inverse reinforcement learning.
\newblock In \emph{International Conference on Machine Learning (ICML)}, 2004.

\bibitem[Aghasadeghi \& Bretl(2011)Aghasadeghi and Bretl]{ab-meirlpi-11}
Aghasadeghi, N. and Bretl, T.
\newblock Maximum entropy inverse reinforcement learning in continuous state
  spaces with path integrals.
\newblock In \emph{International Conference on Intelligent Robots and Systems
  (IROS)}, 2011.

\bibitem[Audiffren et~al.(2015)Audiffren, Valko, Lazaric, and
  Ghavamzadeh]{avlg-ssirl-15}
Audiffren, J., Valko, M., Lazaric, A., and Ghavamzadeh, M.
\newblock {Maximum Entropy Semi-Supervised Inverse Reinforcement Learning}.
\newblock In \emph{{International Joint Conference on Artificial Intelligence
  (IJCAI)}}, July 2015.

\bibitem[Bagnell \& Schneider(2003)Bagnell and Schneider]{bagnell2003covariant}
Bagnell, J.~A. and Schneider, J.
\newblock Covariant policy search.
\newblock In \emph{International Joint Conference on Artificial Intelligence
  (IJCAI)}, 2003.

\bibitem[Boularias et~al.(2011)Boularias, Kober, and Peters]{bkp-reirl-11}
Boularias, A., Kober, J., and Peters, J.
\newblock Relative entropy inverse reinforcement learning.
\newblock In \emph{International Conference on Artificial Intelligence and
  Statistics (AISTATS)}, 2011.

\bibitem[Byravan et~al.(2015)Byravan, Monfort, Ziebart, Boots, and
  Fox]{bmzbf-gbioc-15}
Byravan, A., Monfort, M., Ziebart, B., Boots, B., and Fox, D.
\newblock Graph-based inverse optimal control for robot manipulation.
\newblock In \emph{International Joint Conference on Artificial Intelligence
  (IJCAI)}, 2015.

\bibitem[Doerr et~al.(2015)Doerr, Ratliff, Bohg, Toussaint, and
  Schaal]{drbts-dlmioc-15}
Doerr, A., Ratliff, N., Bohg, J., Toussaint, M., and Schaal, S.
\newblock Direct loss minimization inverse optimal control.
\newblock In \emph{Proceedings of Robotics: Science and Systems (R:SS)}, Rome,
  Italy, July 2015.

\bibitem[Dragan \& Srinivasa(2012)Dragan and Srinivasa]{ds-fat-12}
Dragan, Anca and Srinivasa, Siddhartha.
\newblock Formalizing assistive teleoperation.
\newblock In \emph{Proceedings of Robotics: Science and Systems (R:SS)},
  Sydney, Australia, July 2012.

\bibitem[Finn et~al.(2016)Finn, Tan, Duan, Darrell, Levine, and
  Abbeel]{ftddla-dsae-15}
Finn, Chelsea, Tan, Xin~Yu, Duan, Yan, Darrell, Trevor, Levine, Sergey, and
  Abbeel, Pieter.
\newblock Deep spatial autoencoders for visuomotor learning.
\newblock \emph{International Conference on Robotics and Automation (ICRA)},
  2016.

\bibitem[Huang \& Kitani(2014)Huang and Kitani]{hk-arfdh-14}
Huang, D. and Kitani, K.
\newblock Action-reaction: Forecasting the dynamics of human interaction.
\newblock In \emph{European Conference on Computer Vision (ECCV)}, 2014.

\bibitem[Kalakrishnan et~al.(2013)Kalakrishnan, Pastor, Righetti, and
  Schaal]{kprs-lofm-13}
Kalakrishnan, M., Pastor, P., Righetti, L., and Schaal, S.
\newblock Learning objective functions for manipulation.
\newblock In \emph{International Conference on Robotics and Automation (ICRA)},
  2013.

\bibitem[Levine \& Abbeel(2014)Levine and Abbeel]{la-lnnpg-14}
Levine, S. and Abbeel, P.
\newblock Learning neural network policies with guided policy search under
  unknown dynamics.
\newblock In \emph{Advances in Neural Information Processing Systems (NIPS)},
  2014.

\bibitem[Levine \& Koltun(2012)Levine and Koltun]{lk-cioc-12}
Levine, S. and Koltun, V.
\newblock Continuous inverse optimal control with locally optimal examples.
\newblock In \emph{International Conference on Machine Learning (ICML)}, 2012.

\bibitem[Levine et~al.(2011)Levine, Popovic, and Koltun]{lpk-gpirl-11}
Levine, S., Popovic, Z., and Koltun, V.
\newblock Nonlinear inverse reinforcement learning with gaussian processes.
\newblock In \emph{Advances in Neural Information Processing Systems (NIPS)},
  2011.

\bibitem[Levine et~al.(2015)Levine, Wagener, and Abbeel]{lwa-lnnpg-15}
Levine, S., Wagener, N., and Abbeel, P.
\newblock Learning contact-rich manipulation skills with guided policy search.
\newblock In \emph{International Conference on Robotics and Automation (ICRA)},
  2015.

\bibitem[Monfort et~al.(2015)Monfort, Lake, Ziebart, Lucey, and
  Tenenbaum]{mlzlt-shgpi-15}
Monfort, M., Lake, B.~M., Ziebart, B., Lucey, P., and Tenenbaum, J.
\newblock Softstar: Heuristic-guided probabilistic inference.
\newblock In \emph{Advances in Neural Information Processing Systems}, pp.\
  2746--2754, 2015.

\bibitem[Muelling et~al.(2014)Muelling, Boularias, Mohler, Sch{\"o}lkopf, and
  Peters]{mbmsp-ttirl-14}
Muelling, K., Boularias, A., Mohler, B., Sch{\"o}lkopf, B., and Peters, J.
\newblock Learning strategies in table tennis using inverse reinforcement
  learning.
\newblock \emph{Biological Cybernetics}, 108\penalty0 (5), 2014.

\bibitem[Ng et~al.(1999)Ng, Harada, and Russell]{nhr-tars-99}
Ng, A., Harada, D., and Russell, S.
\newblock Policy invariance under reward transformations: Theory and
  application to reward shaping.
\newblock In \emph{International Conference on Machine Learning (ICML)}, 1999.

\bibitem[Ng et~al.(2000)Ng, Russell, et~al.]{nr-airl-00}
Ng, A., Russell, S., et~al.
\newblock Algorithms for inverse reinforcement learning.
\newblock In \emph{International Conference on Machine Learning (ICML)}, 2000.

\bibitem[Peters et~al.(2010)Peters, M{\"u}lling, and Alt{\"u}n]{pma-reps-10}
Peters, J., M{\"u}lling, K., and Alt{\"u}n, Y.
\newblock Relative entropy policy search.
\newblock In \emph{AAAI Conference on Artificial Intelligence}, 2010.

\bibitem[Ramachandran \& Amir(2007)Ramachandran and Amir]{ra-birl-07}
Ramachandran, D. and Amir, E.
\newblock Bayesian inverse reinforcement learning.
\newblock In \emph{AAAI Conference on Artificial Intelligence}, volume~51,
  2007.

\bibitem[Ratliff et~al.(2006)Ratliff, Bagnell, and Zinkevich]{rbz-mmp-06}
Ratliff, N., Bagnell, J.~A., and Zinkevich, M.~A.
\newblock Maximum margin planning.
\newblock In \emph{International Conference on Machine Learning (ICML)}, 2006.

\bibitem[Ratliff et~al.(2007)Ratliff, Bradley, Bagnell, and
  Chestnutt]{rbbc-bsp-07}
Ratliff, N., Bradley, D., Bagnell, J.~A., and Chestnutt, J.
\newblock Boosting structured prediction for imitation learning.
\newblock 2007.

\bibitem[Ratliff et~al.(2009)Ratliff, Silver, and Bagnell]{rsb-learch-09}
Ratliff, N., Silver, D., and Bagnell, J.~A.
\newblock Learning to search: Functional gradient techniques for imitation
  learning.
\newblock \emph{Autonomous Robots}, 27\penalty0 (1), 2009.

\bibitem[Rawlik \& Vijayakumar(2013)Rawlik and
  Vijayakumar]{rawlik2013stochastic}
Rawlik, K. and Vijayakumar, S.
\newblock On stochastic optimal control and reinforcement learning by
  approximate inference.
\newblock \emph{Robotics}, 2013.

\bibitem[Todorov(2006)]{t-lsmdp-06}
Todorov, E.
\newblock Linearly-solvable markov decision problems.
\newblock In \emph{Advances in Neural Information Processing Systems (NIPS)},
  2006.

\bibitem[Todorov et~al.(2012)Todorov, Erez, and Tassa]{tet-mjc-12}
Todorov, E., Erez, T., and Tassa, Y.
\newblock {MuJoCo}: A physics engine for model-based control.
\newblock In \emph{International Conference on Intelligent Robots and Systems
  (IROS)}, 2012.

\bibitem[Tzeng et~al.(2015)Tzeng, Hoffman, Darrell, and Saenko]{thds-dtad-15}
Tzeng, E., Hoffman, J., Darrell, T., and Saenko, K.
\newblock Simultaneous deep transfer across domains and tasks.
\newblock In \emph{International Conference on Computer Vision (ICCV)}, 2015.

\bibitem[Wulfmeier et~al.(2015)Wulfmeier, Ondruska, and Posner]{wop-dirl-15}
Wulfmeier, M., Ondruska, P., and Posner, I.
\newblock Maximum entropy deep inverse reinforcement learning.
\newblock \emph{arXiv preprint arXiv:1507.04888}, 2015.

\bibitem[Ziebart(2010)]{z-mpabp-10}
Ziebart, B.
\newblock \emph{Modeling purposeful adaptive behavior with the principle of
  maximum causal entropy}.
\newblock PhD thesis, Carnegie Mellon University, 2010.

\bibitem[Ziebart et~al.(2008)Ziebart, Maas, Bagnell, and Dey]{zmbd-meirl-08}
Ziebart, B., Maas, A., Bagnell, J.~A., and Dey, A.~K.
\newblock Maximum entropy inverse reinforcement learning.
\newblock In \emph{AAAI Conference on Artificial Intelligence}, 2008.

\end{thebibliography}
\bibliographystyle{icml2016}

\clearpage

\appendix

\section{Policy Optimization under Unknown Dynamics}
\label{app:trajopt}

The policy optimization procedure employed in this work follows the method described by Levine and Abbeel~\cite{la-lnnpg-14}, which we summarize in this appendix. The aim is to optimize Gaussian trajectory distributions $q(\traj) = q(\state_1)\prod_t q(\state_{t+1}|\st,\at)q(\at|\st)$ with respect to their expected cost $E_{q(\traj)}[\cost_\params(\traj)]$. This optimization can be performed by iteratively optimizing $E_{q(\traj)}[\cost_\params(\traj)]$ with respect to the linear-Gaussian conditionals $q(\at|\st)$ under a linear-Gaussian estimate of the dynamics $q(\state_{t+1}|\st,\at)$. This optimization can be performed using the standard linear-quadratic regulator (LQR). However, when the dynamics of the system are not known, the linearization $q(\state_{t+1}|\st,\at)$ cannot be obtained directly. Instead, Levine and Abbeel~\cite{la-lnnpg-14} propose to estimate the local linear-Gaussian dynamics $q(\state_{t+1}|\st,\at)$ using samples from $q(\traj)$, which can be obtained by running the linear-Gaussian controller $q(\at|\st)$ on the physical system. The policy optimization procedure then consists of iteratively generating samples from $q(\at|\st)$, fitting $q(\state_{t+1}|\st,\at)$ to these samples, and updating $q(\at|\st)$ under these fitted dynamics by using LQR.

This policy optimization procedure has several important nuances. First, the LQR update can modify the controller $q(\at|\st)$ arbitrarily far from the previous controller. However, because the real dynamics are not linear, this new controller might experience very different dynamics on the physical system than the linear-Gaussian dynamics $q(\state_{t+1}|\st,\at)$ used for the update. To limit the change in the dynamics under the current controller, Levine and Abbeel~\cite{la-lnnpg-14} propose solving a modified, constrained problem for updating $q(\at|\st)$, given as following:
\[
\max_{q \in \gauss} E_q[\cost_\params(\traj)] \text{ s.t. } \kl(q \| \hat{q}) \leq \epsilon,
\]
\noindent where $\hat{q}$ is the previous controller. This constrained problem finds a new trajectory distribution $q(\traj)$ that is close to the previous distribution $\hat{q}(\traj)$, so that the dynamics violation is not too severe. The step size $\epsilon$ can be chosen adaptively based on the degree to which the linear-Gaussian dynamics are successful in predicting the current cost \cite{lwa-lnnpg-15}. Note that when policy optimization is interleaved with IOC, special care must be taken when adapting this step size. We found that an effective strategy was to use the step size rule described in prior work \cite{lwa-lnnpg-15}. This update involves comparing the predicted and actual improvement in the cost. We used the preceding cost function to measure both improvements since this cost was used to make the update.


The second nuance in this procedure is in the scheme used to estimate the dynamics $q(\state_{t+1}|\st,\at)$. Since these dynamics are linear-Gaussian, they can be estimated by solving a separate linear regression problem at each time step, using the samples gathered at this iteration. The sample complexity of this procedure scales linearly with the dimensionality of the system. However, this sample complexity can be reduced dramatically if we consider the fact that they dynamics at nearby time steps are strongly correlated, even across iterations (due to the previously mentioned KL-divergence constraint). This property can be exploited by fitting a crude global model to all of the samples gathered during the policy optimization procedure, and then using this global model as a prior for the linear regression. A good choice for this global model is a Gaussian mixture model (GMM) over tuples of the form $(\st,\at,\state_{t+1})$, as discussed in prior work \cite{la-lnnpg-14}. This GMM is refitted at each iteration using all available interaction data, and acts as a prior when fitting the time-varying linear-Gaussian dynamics $q(\state_{t+1}|\st,\at)$.

\section{Consistency Evaluation}
\label{app:consistent}

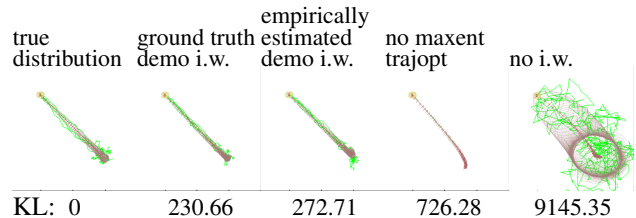
\begin{figure}
\setlength{\unitlength}{0.5\columnwidth}
\begin{picture}(1.0,0.75) \linethickness{0.5pt}

\put(0.0,0.1){\includegraphics[height=0.2\columnwidth]{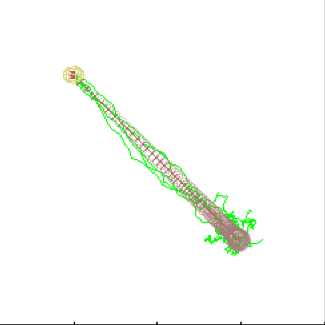}}
\put(0.4,0.1){\includegraphics[height=0.2\columnwidth]{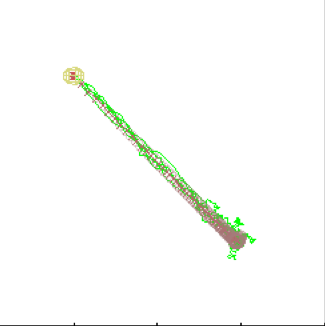}}
\put(0.8,0.1){\includegraphics[height=0.2\columnwidth]{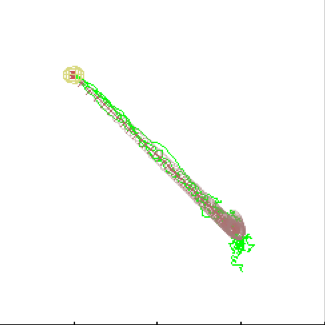}}
\put(1.2,0.1){\includegraphics[height=0.2\columnwidth]{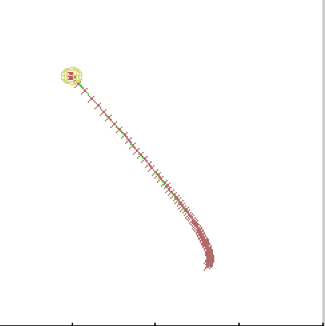}}
\put(1.6,0.1){\includegraphics[height=0.2\columnwidth]{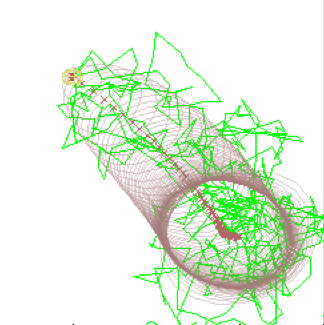}}

\put(0.0,0.57){\footnotesize{true}}
\put(0.0,0.5){\footnotesize{distribution}}
\put(0.4,0.57){\footnotesize{ground truth}}
\put(0.4,0.5){\footnotesize{demo i.w.}}
\put(0.8,0.64){\footnotesize{empirically}}
\put(0.8,0.57){\footnotesize{estimated}}
\put(0.8,0.5){\footnotesize{demo i.w.}}
\put(1.2,0.57){\footnotesize{no maxent}}
\put(1.2,0.5){\footnotesize{trajopt}}
\put(1.6,0.5){\footnotesize{no i.w.}}

\put(0.0,0.02){KL:}
\put(0.18,0.02){\footnotesize{0}}
\put(0.5,0.02){\footnotesize{230.66}}
\put(0.9,0.02){\footnotesize{272.71}}
\put(1.3,0.02){\footnotesize{726.28}}
\put(1.68,0.02){\footnotesize{9145.35}}

\end{picture}
\vspace{-0.07in}
\caption{KL divergence between trajectories produced by our method, and various ablations, to the true distribution. Guided cost learning recovers trajectories that come close to both the mean and variance of the true distribution using 40 demonstrated trajectories, whereas the algorithm without MaxEnt policy optimization or without importance weights recovers the mean but not the variance.
\label{fig:pointmass}
}
\end{figure}

We evaluated the consistency of our algorithm by generating 40 demonstrations from 4 known linear Gaussian trajectory distributions of a second order point mass system, each traveling to the origin from different starting positions. The purpose of this experiment is to verify that, in simple domains where the exact cost function can be learned, our method is able to recover the true cost function successfully. To do so, we measured the KL divergence between the trajectories produced by our method and the true distribution underlying the set of demonstrations.
As shown in Figure~\ref{fig:pointmass}, the trajectory distribution produced by guided cost learning with ground truth demo importance weights (weights based on the true distribution from which the demonstrations were sampled, which is generally unknown) comes very close to the true distribution, with a KL divergence of $230.66$ summed over 100 timesteps. Empirically estimating the importance weights of the demos produces trajectories with a slightly higher KL divergence of $272.71$, costing us very little in this domain. Dropping the demo and sample importance weights entirely recovers a similar mean, but significantly overestimates the variance. Finally, running the algorithm without a maximum entropy term in the policy optimization objective (see Section~\ref{sec:trajopt}) produces trajectories with similar mean, but 0 variance. These results indicate that correctly incorporating importance weights into sample-based maximum entropy IOC is crucial for recovering the right cost function. This contrasts with prior work, which suggests dropping the importance weights \cite{kprs-lofm-13}.

\section{Neural Network Parametrization and Initialization}
\label{app:nn}

We use expressive neural network function approximators to represent the cost, using the form:
\begin{equation*}
\cost_\params(\st,\at) = \| A \mathbf{y}_t + b \|^2 + w_\action \| \at \|^2
\label{eqn:costnn}
\end{equation*}

This parametrization can be viewed as a cost that is quadratic in a set of learned nonlinear features $\mathbf{y}_t = f_\params(\st)$ where $f_\params$ is a multilayer neural network with rectifying nonlinearities of the form $\max(z,0)$. Since simpler cost functions are generally preferred, we initialize the $f_\params$ to be the identity function by setting the parameters of the first fully-connected layer to contain the identity matrix and the negative identity matrix (producing hidden units which are double the dimension of the input), and all subsequent layers to the identity matrix. We found that this initialization improved generalization of the learned cost.

\section{Detailed Description of Task Setup}
\label{app:tasks}

All of the simulated experiments used the MuJoCo simulation package~\citep{tet-mjc-12}, with simulated frictional contacts and torque motors at the joints used for actuation. All of the real world experiments were on a PR2 robot, using its 7 DOF arm controlled via direct effort control. Both the simulated and real world controllers were run for 5 seconds at 20 Hz resulting in 100 time steps per rollout. We describe the details of each system below.

In all tasks except for 2D navigation (which has a small state space and complex cost), we chose the dimension of the hidden layers to be approximately double the size of the state, making it capable of representing the identity function.

\paragraph{2D Navigation:} The 2D navigation task has 4 state dimensions (2D position and velocity) and 2 action dimensions. Forty demonstrations were generated by optimizing trajectories for 32 randomly selected positions, with at least 1 demonstration from each starting position. The neural network cost was parametrized with 2 hidden layers of dimension 40 and a final feature dimension of 20.
\paragraph{Reaching:} The 2D reaching task has 10 dimensions (3 joint angles and velocities, 2-dimensional end effector position and velocity). Twenty demonstrations were generated by optimizing trajectories from 12 different initial states with arbitrarily chosen joint angles. The neural network was parametrized with 2 hidden layers of dimension 24 and a final feature dimension of 100.
\paragraph{Peg insertion:} The 3D peg insertion task has 26 dimensions (7 joint angles, the pose of 2 points on the peg in 3D, and the velocities of both). Demonstrations were generated by shifting the hole within a 0.1 m $\times$ 0.1 m region on the table.
Twenty demonstrations were generated from sixteen demonstration conditions. The neural network was parametrized with 2 hidden layers of dimension 52 and a final feature dimension of 100.
\paragraph{Dish:} The dish placing task has 32 dimensions (7 joint angles, the 3D position of 3 points on the end effector, and the velocities of both). Twenty demonstrations were collected via kinesthetic teaching on nine positions along a 43 cm dish rack. A tenth position, spatially located within the demonstrated positions, was used during IOC.
The input to the cost consisted of the 3 end effector points in 3D relative to the target pose (which fully define the pose of the gripper) and their velocities. The neural network was parametrized with 1 hidden layer of dimension 64 and a final feature dimension of 100. Success was based on whether or not the plate was in the correct slot and not broken.
\paragraph{Pouring:} The pouring task has has 40 dimensions (7 joint angles and velocities, the 3D position of 3 points on the end effector and their velocities, 2 learned visual feature points in 2D and their velocities). Thirty demonstrations were collected via kinesthetic teaching. For each demonstration, the target cup was placed at a different position on the table within a 28 cm $\times$ 13 cm rectangle. The autoencoder was trained on images from the 30 demonstrations (consisting of 3000 images total). The input to the cost was the same as the state but omitting the joint angles and velocities.
The neural network was parametrized with 1 hidden layer of dimension 80 and a final feature dimension of 100. To measure success, we placed 15 almonds in the grasped cup and measured the percentage of the almonds that were in the target cup after the executed motion.

\section{Regularization Evaluation}
\label{app:ablation}

\begin{figure}
\setlength{\unitlength}{0.5\columnwidth}
\begin{picture}(1.0,1.9) \linethickness{0.5pt}

\put(0.05,0.88){\includegraphics[height=0.48\columnwidth]{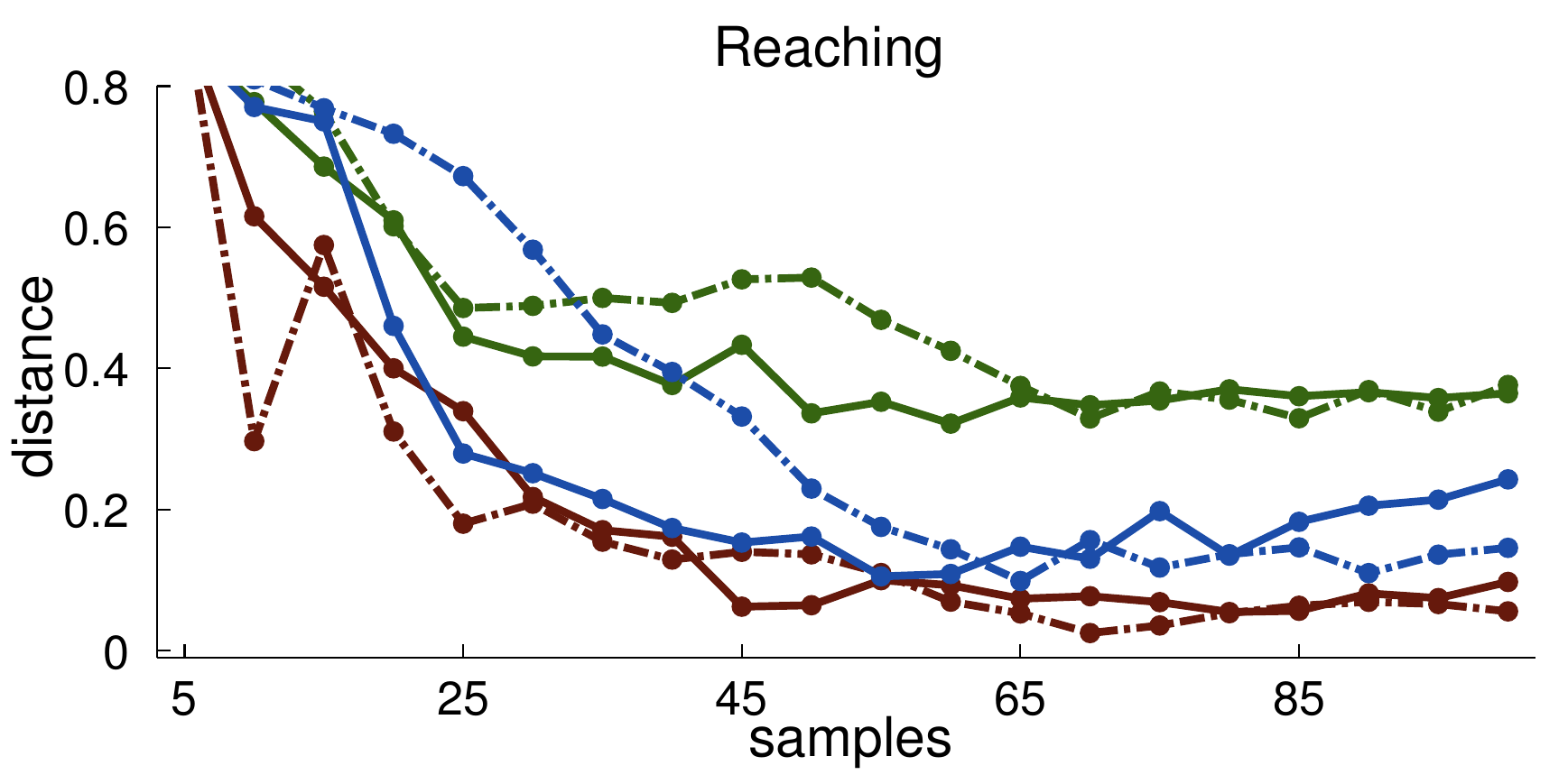}}

\put(0.05,-0.08){\includegraphics[height=0.48\columnwidth]{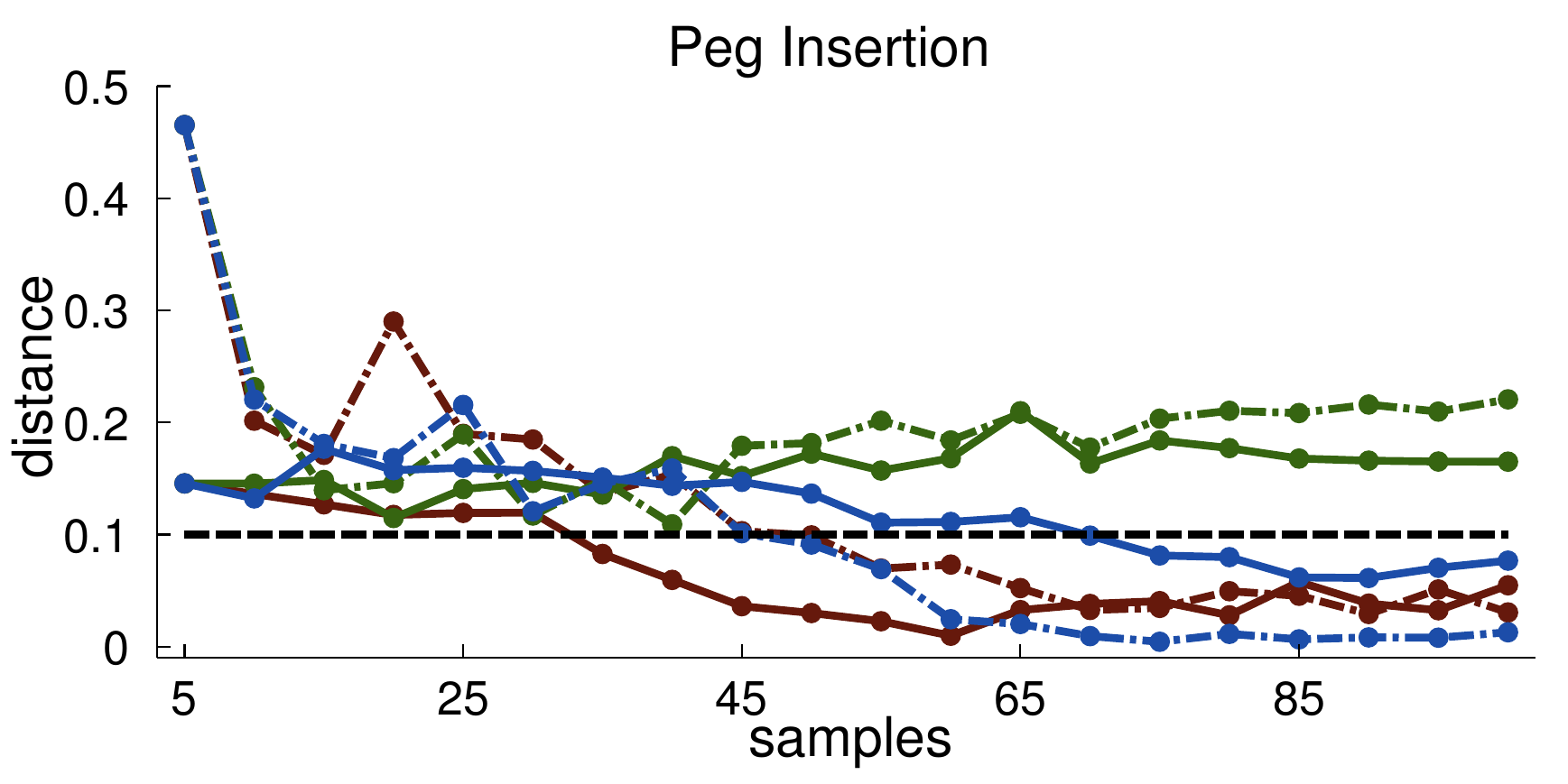}}
\put(1.34,1.475){\includegraphics[height=0.18\columnwidth]{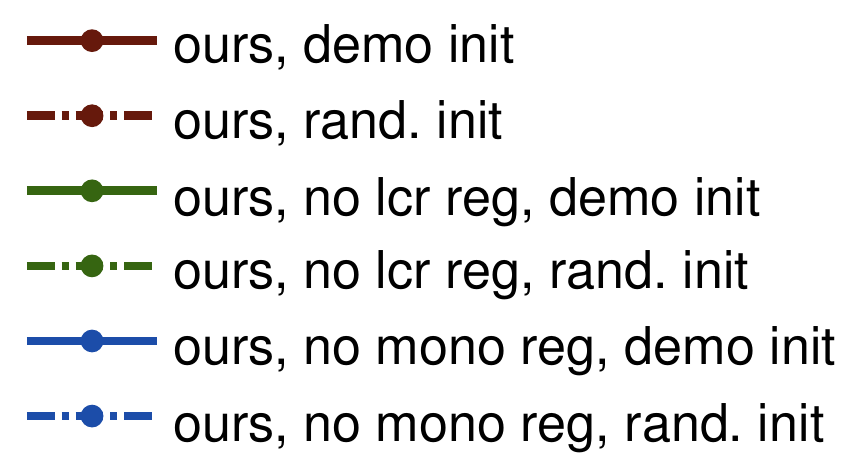}}
\end{picture}
\vspace{0.02in}
\caption{Comparison showing ablations of our method with leaving out one of the two regularization terms. The monotonic regularization improves performance in three of the four task settings, and the local constant rate regularization significantly improves performance in all settings. Reported distance is averaged over four runs of IOC on four different initial conditions.
\label{fig:ablation}
\vspace{-0.15in}
}
\end{figure}

We evaluated the performance with and without each of the two regularization terms proposed in Section~\ref{sec:regularization} on the simulated reaching and peg insertion tasks. As shown in Figure~\ref{fig:ablation}, both regularization terms help performance. Notably, the learned trajectories fail to insert the peg into the hole when the cost is learned using no local constant rate regularization.

\end{document}